\pdfoutput=1

\documentclass[11pt]{article}

\PassOptionsToPackage{dvipsnames,table}{xcolor}

\usepackage[final]{acl}

\usepackage{times}
\usepackage{latexsym}

\usepackage[T1]{fontenc}

\usepackage[utf8]{inputenc}

\usepackage{microtype}

\usepackage{inconsolata}

\usepackage{graphicx}

\usepackage{amsmath,amsfonts,bm,mathtools,amssymb}
\usepackage{booktabs}
\usepackage{graphicx}
\usepackage{multirow,longtable}
\usepackage{nccmath}
\usepackage{makecell}
\usepackage{footmisc}
\usepackage{cleveref}
\usepackage{colortbl}
\usepackage{subcaption}  %
\usepackage{multirow, makecell}
\Crefname{section}{\S}{\S\S}

\usepackage{xspace}

\title{Dual Alignment Between Language Model Layers \\ and Human Sentence Processing}

\author{Tatsuki Kuribayashi${}^{1,2}$\, Alex Warstadt${}^{3}$\, Yohei Oseki${}^{4}$  Ethan Gotlieb Wilcox${}^{5}$ \\
        ${}^{1}$MBZUAI \,
        ${}^{2}$Tohoku University\,
        ${}^{3}$UC San Diego \, \\
        ${}^{4}$The University of Tokyo \,
        ${}^{5}$Georgetown University \\
  \texttt{tatsuki.kuribayashi@mbzuai.ac.ae} $\;\;\;\;$
  \texttt{awarstadt@ucsd.edu} \\
  \texttt{oseki@g.ecc.u-tokyo.ac.jp}   $\;\;\;\;$
    \texttt{ethan.wilcox@georgetown.edu}
}

\newcommand{\mymathcolor}{MidnightBlue}
\newcommand{\orange}{Bittersweet}
\newcommand{\timestep}{\textcolor{\mymathcolor}{\ensuremath{t}}\xspace}
\newcommand{\word}{\textcolor{\mymathcolor}{\ensuremath{w_{\timestep}}}\xspace}

\newcommand{\prevwords}{\textcolor{\mymathcolor}{\ensuremath{\mathbf{w}_{<\timestep}}}\xspace}

\newcommand{\defn}[1]{\textbf{#1}\xspace}
\newcommand{\Word}{\textcolor{\mymathcolor}{\ensuremath{W}}\xspace}

\newcommand{\layer}{\textcolor{\orange}{\ensuremath{l}}\xspace}
\newcommand{\prevlayer}{\textcolor{\orange}{\ensuremath{l-1}}\xspace}

\newcommand{\w}{\textcolor{\mymathcolor}{\ensuremath{w}}\xspace}

\begin{document}
\maketitle
\begin{abstract}
A recent study~\cite{kuribayashi2025largelanguagemodelshumanlike} has shown that human sentence processing behavior, typically measured on syntactically unchallenging constructions, can be effectively modeled using surprisal from early layers of large language models (LLMs). 
This raises the question of whether such advantages of internal layers extend to more syntactically challenging constructions, where surprisal has been reported to underestimate human cognitive effort.
In this paper, we begin by exploring internal layers that better estimate human cognitive effort observed in syntactic ambiguity processing in English.
Our experiments show that, in contrast to naturalistic reading, later layers better estimate such a cognitive effort, but still underestimate the human data. 
This \textit{dual alignment} sheds light on different modes of sentence processing in humans and LMs: naturalistic reading employs a somewhat weak prediction akin to earlier layers of LMs, while syntactically challenging processing requires more fully-contextualized representations, better modeled by later layers of LMs.
Motivated by these findings, we also explore several probability-update measures using shallow and deep layers of LMs, showing a complementary advantage to single-layer's surprisal in reading time modeling.
\newline
\newline
\hspace{.1em}\includegraphics[width=1.05em,height=1.05em]{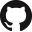}\hspace{.5em}\parbox{\dimexpr\linewidth-2\fboxsep-2\fboxrule}
  {\small \url{https://github.com/kuribayashi4/internal_surprisal_targeted_assessment}}
\end{abstract}

\section{Introduction}
\label{sec:intro}
A central goal in computational psycholinguistics is to understand human sentence processing through constructing a computational model that can simulate it~\cite{Crocker2010-cp}. 
Language models (LMs) have offered a framework to explore the candidates for cognitively plausible models, motivated by the widely held view that \textit{prediction} is a core principle of human sentence processing~\cite{Clark2013-xs,Levy2008Expectation-basedComprehension,Smith2013-ap}.
As LMs are fundamentally designed as next-word prediction machines, they have served as a tool to estimate the predictability of words and contributed to exploring the role of prediction in human language, particularly in online sentence processing (\citealt{frank2011insensitivity,Goodkind2018PredictiveQuality,Hale2018FindingSearch,Wilcox2020OnBehavior,Oh2023-zw,kuribayashi2025largelanguagemodelshumanlike}; i.a.).

\begin{figure}[t]
    \centering
    \includegraphics[width=0.48\textwidth]{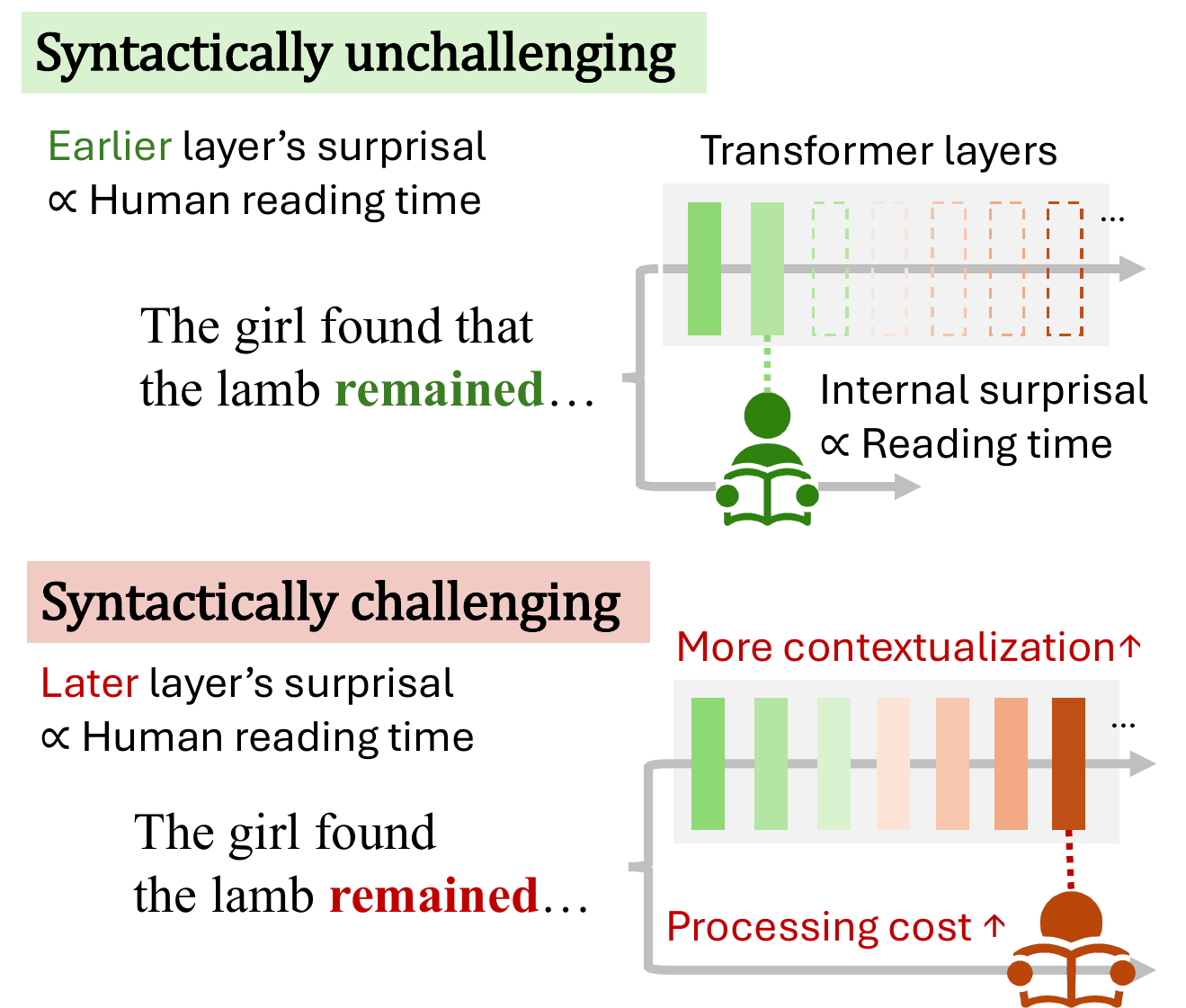}
    \caption{We examine surprisal from internal layers of Transformer LMs to better capture human sentence processing behavior and find that deeper layers align better in syntactically challenging constructions.}
    \label{fig:fig1}
\end{figure}

Existing studies have demonstrated both the successes and limitations of accurate predictability estimation by modern LMs in cognitive modeling. 
In particular, surprisal, defined as $-\log p(\word\mid\prevwords)$ for a word \word and a context of previous words \prevwords\ and estimated from some LMs, has proven to be a strong predictor of human reading behavior~\cite{frank2011insensitivity,Goodkind2018PredictiveQuality,Wilcox2023-pi}. 
However, surprisal from very large LMs, despite arguably better alignment with ground truth text distribution, has been shown empirically to misalign with human naturalistic reading behaviors throughout an entire corpus~\cite{kuribayashi-etal-2021-lower,Oh2023-zw,Shain2022-qv,de-varda-marelli-2023-scaling,Boeve2025-yi}.
We refer to this type of shortcoming as \defn{holistic misalignment}.
At the same time, surprisal from all LMs so far tested has been found to underestimate the cognitive load associated with syntactically challenging constructions, such as garden-path sentences or ungrammatical sentence regions~\cite{Van_Schijndel2021-sm,Wilcox2021-gy,Arehalli2022-nb,Huang2024-qe,Timkey2025-cn}.
We refer to this as \defn{targeted misalignment}.
Targeted misalignment has been found in contexts where humans experience high cognitive load and exhibit a substantial slowdown in reading, the magnitude of which is not reflected in models' surprisal values.
A recent study~\cite{kuribayashi2025largelanguagemodelshumanlike} addressed the holistic misalignment issue by showing that surprisal decoded from earlier layers of LMs, rather than final layers, better matches human-like reading behavior on syntactically unchallenging, naturalistic corpora.

We ask whether targeted misalignment can be reconciled by using surprisal from internal model layers (\cref{sec:exp1} and \cref{sec:exp2}).
Our experimental results demonstrate that, in contrast to the naturalistic reading results, earlier layers do not better simulate the contrastive human reading slowdown in syntactically challenging contexts. 
They compute almost the same surprisal in both syntactically ambiguous and unambiguous conditions, reflecting an overly severe recency bias and syntactic insensitivity.
Our results, therefore, contrast with those presented in \citet{kuribayashi2025largelanguagemodelshumanlike}; earlier layers alone are not a cognitively plausible model of human sentence processing when extended to syntactically challenging contexts.

Widening our investigation to the whole model, we find that later layers align better with syntactic ambiguity processing behavior.
However, they still produce an \emph{underestimate} of the human reading data.
This \textit{dual alignment} (Figure~\ref{fig:fig1}) between LM layers and human sentence processing stages suggests that different stages of human sentence processing may correspond to different layers of LMs; in particular, normal naturalistic processing can be aligned with earlier layers' prediction, while slower, late-stage processing (e.g., reanalysis) demands later layers' more contextualized representations.
This partially supports the recent proposed correspondence between LMs' forward computation to human language processing stages~\cite{Tenney2019-bb,hu2025signatures,kuribayashi2025largelanguagemodelshumanlike}.

Combining our two empirical findings, we propose using contrastive surprisal from early vs. late layers as a measure of the \emph{degree of belief update} between shallow and fully-contextualized processing. We hypothesize that this measure can be used to identify data points that will be contextually demanding for humans to process (\cref{sec:exp3}). 
This is based on the observation that, generally, syntactically challenging constructions incurred greater qualitative change in surprisal across layers.
We exemplify this proposal by showing the effectiveness of surprisal update as a predictor in reading time modeling, but leave a full analysis as a direction for future research.

\section{Background}
\subsection{Surprisal theory}
Humans exhibit different cognitive load (e.g., measured by reading time) for different interest areas (e.g., words or tokens) in a text during reading.
Surprisal has proven to be a robust predictor of reading time across languages and experimental paradigms~\cite{Levy2008Expectation-basedComprehension,Demberg2008-fd, Wilcox2023-pi}. 
The surprisal~\cite{Cover1999-up} of a word $\word \in \Word$ in context $\prevwords := [w_0, \cdots, w_{t-1}]^\top$ is defined as $-\log P_t(\Word=\word|\prevwords)$, where $P_t: W \to [0,1]$ is a family of conditional distributions assigning a probability to a word $w$ at time step $t$ given its prefix.
Thus, the more unexpected \word is, the more costly it is for humans to process.
Empirically, this cost has been found to scale linearly with its negative log probability~\cite{Smith2013-ap,Shain2022-qv}.

\begin{table*}[t]
\centering
\small
\tabcolsep=2pt
\begin{tabular}{cl}
\toprule
\textbf{Phenomena} & \textbf{Example} \\
\midrule

\multirow{2}{*}{MVRR} 
 & $D^+$: The girl fed the lamb \textbf{\textit{remained}} relatively calm before the sunset in silence. \\
 & $D^-$: The girl \underline{who was} fed the lamb \textbf{\textit{remained}} relatively calm before the sunset in silence. \\

\midrule
\multirow{2}{*}{NPS} 
 & $D^+$: The girl found the lamb \textbf{\textit{remained}} relatively calm near the wooden fence. \\
 & $D^-$: The girl found \underline{that} the lamb \textbf{\textit{remained}} relatively calm near the wooden fence. \\

\midrule
\multirow{2}{*}{NPZ} 
 & $D^+$: When the girl attacked the lamb \textbf{\textit{remained}} relatively calm despite the sudden noise. \\
 & $D^-$: When the girl attacked\underline{,} the lamb \textbf{\textit{remained}} relatively calm despite the sudden noise. \\

\midrule
\multirow{2}{*}{RC} 
 & $D^+$: The bus driver that \textbf{\textit{the}} kids followed waited patiently at dawn. \\
 & $D^-$: The bus driver that \textbf{\textit{followed}} the kids waited patiently at dawn. \\

\midrule
\multirow{2}{*}{Attachment} 
 & $D^+$: Janet charmed the executive of the assistant\underline{s} who \textbf{\textit{decides}} almost everything during long weekly meetings. \\
 & $D^-$: Janet charmed the executive\underline{s} of the assistant who \textbf{\textit{decides}} almost everything during long weekly meetings. \\

\bottomrule
\end{tabular}
\caption{Examples of pairs of syntactically ambiguous ($\in D^+$) and unambiguous ($\in D^-$) sentences. The underlined parts are minimal differences between the two conditions. The bold part is the disambiguating point $t^*$, as well as the first word of the region of interest (\textsc{RoI}). Examples are borrowed from~\citet{Huang2024-qe}.}
\label{tab:examples}
\end{table*}

\subsection{Internal surprisal from Transformers}
\label{subsec:internal_surprisal}

To review the decoder-based Transformer architecture, the model consists of a stack of layers that parameterize the anticipation of the next word $P(\cdot\mid\prevwords)$ by iteratively processing the context through self-attention.
Specifically, in each layer \layer for each token $i$, the model integrates the previous layer's representations up to and including $\textcolor{MidnightBlue}{i}$ to output a representation $\bm h_{i}^{(\layer)} \in \mathbb{R}^d$:
\begin{align}
    \label{eq:layers}
    \bm h^{(\layer)}_i &= \mathcal{F}^{(\layer)} \bigl(\bm h^{(\prevlayer)}_0,\dots, \bm h^{(\prevlayer)}_{i}\bigr)\;\;\mathrm{,} \\
    \label{eq:first_layer}
    \bm h^{(0)}_i &= \mathrm{emb}(\textcolor{MidnightBlue}{w_{i}})\;\;\mathrm{,}
\end{align}
\noindent
where $\mathcal{F}^{(\layer)}$ is the forward computation of layer \layer. 
$\bm h^{(0)}_i = \mathrm{emb}(\textcolor{MidnightBlue}{w_{i}}) \in \mathbb{R}^d $ is an input embedding of the word $\textcolor{MidnightBlue}{w_{i}}$.
Layer-specific next word probability is obtained from that layer's representation of the preceding context using logit-lens ~\cite[LLens;][]{logitlens}:

\begin{align}
    \nonumber
    &P^{(\layer)}(\Word=\word\mid\prevwords) = \mathrm{LLens}(\bm h_{t-1}^{(\layer)})^{\texttt{id}(\word)} \\
    &= \mathrm{softmax}(\bm W_U \mathrm{LayerNorm}(\bm h_{t-1}^{(\layer)}))^{\texttt{id}(\word)}\mathrm{,}
    \label{eq:logit-lens}
\end{align}
where $\bm W_U \in \mathbb{R}^{|\mathcal{V}| \times d}$ is an unembedding matrix obtained from the LM's output layer, and $|\mathcal{V}| \in \mathbb{R}$ is the model's vocabulary size. 
The superscript ${\texttt{id}(\word)}$ denotes the element corresponding to word \word in the resulting probability vector.\footnote{When a word is tokenized into multiple subwords, we follow the existing studies~\cite{Oh2023-zw,kuribayashi2025largelanguagemodelshumanlike} to compute the joint probability of the subwords.}
Layer-specific surprisal $S^{(\layer)}_t = -\log P^{(\layer)}(\Word=\word|\prevwords) \in \mathbb{R}_{\geq 0}$ can also be computed.
One limitation with Logit Lense is that it has been empirically found to be less reliable for decoding earlier layers, likely because embeddings do not exist in the same representation space \citep{belrose2023eliciting, langedijk-etal-2024-decoderlens}. We address this limitation in \cref{app:tuned-lens}.
\citet{kuribayashi2025largelanguagemodelshumanlike} explored which layer $\layer$ exhibits a better fit to reading time data. They find that earlier layers typically result in the best prediction for the case of naturalistic reading.

\subsection{Targeted misalignment of surprisal}
\label{subsec:targeted_misalingment}
There are notable cases in which human reading time patterns cannot be explained by (final layer) surprisal, leading to the criticism that surprisal-based predictability alone is insufficient to characterize total processing cost~\cite{Van_Schijndel2021-sm,Arehalli2022-nb,Huang2024-qe,Timkey2025-cn, Wilcox2021-gy}.
Specifically, these studies recorded contrastive reading behavior between two different conditions, where sentences either conform to or violate structural expectations determined by the grammar. While structural expectations can be based on pure grammaticality (as in \citealp{Wilcox2021-gy}), they can also be determined by structural ambiguity. In our study, the test materials consist of pairs of syntactically ambiguous ($\in D^+$) and unambiguous sentences ($\in D^-$).
Examples are given in Table~\ref{tab:examples}.
We choose these phenomena, as managing ambiguity can help us understand the role that context, representations, and prediction play during human language processing.
Previous studies find that the reading time difference between $D^+$ and $D^-$ is underestimated by the magnitude of the surprisal difference between the two conditions.

\begin{figure*}[t]
    \centering
    \begin{subfigure}[b]{\textwidth}
        \centering
        \includegraphics[width=\textwidth]{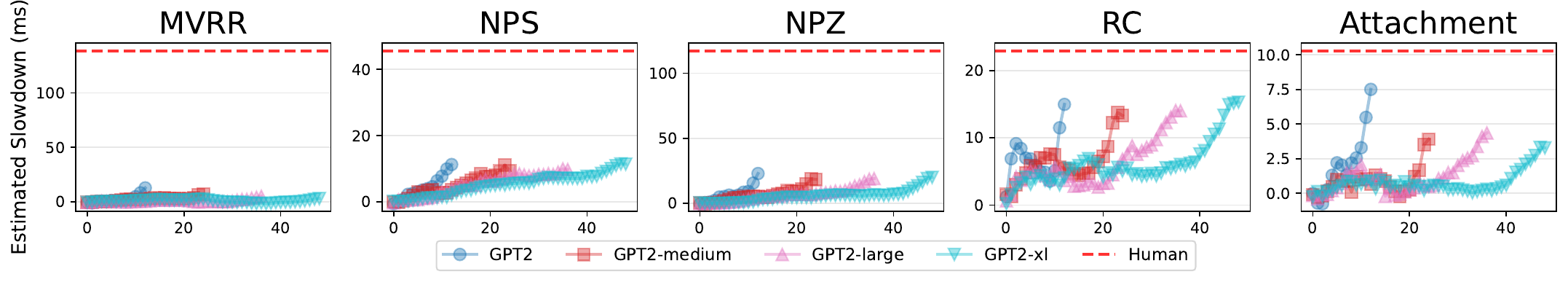}
        \caption{GPT-2 family}
        \label{fig:exp1_gpt2}
    \end{subfigure}
    
    \begin{subfigure}[b]{\textwidth}
        \centering
        \includegraphics[width=\textwidth]{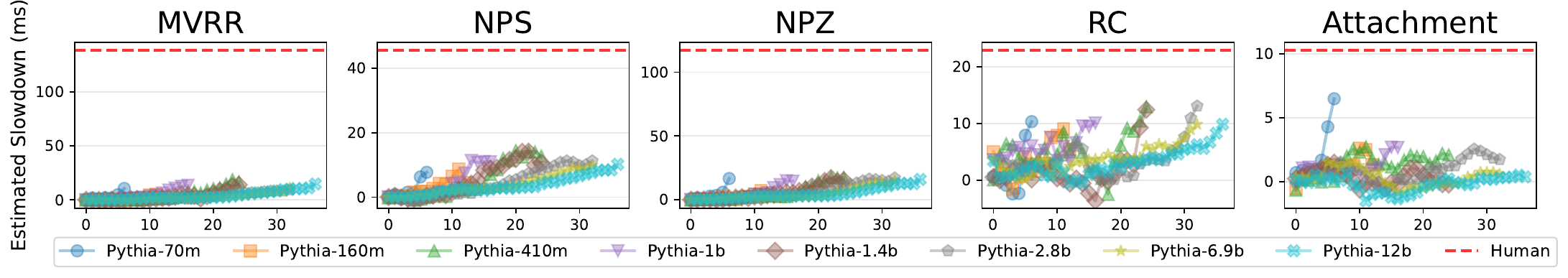}
        \caption{Pythia family}
        \label{fig:exp1_pythia}
    \end{subfigure}
    
    \begin{subfigure}[b]{\textwidth}
        \centering
        \includegraphics[width=\textwidth]{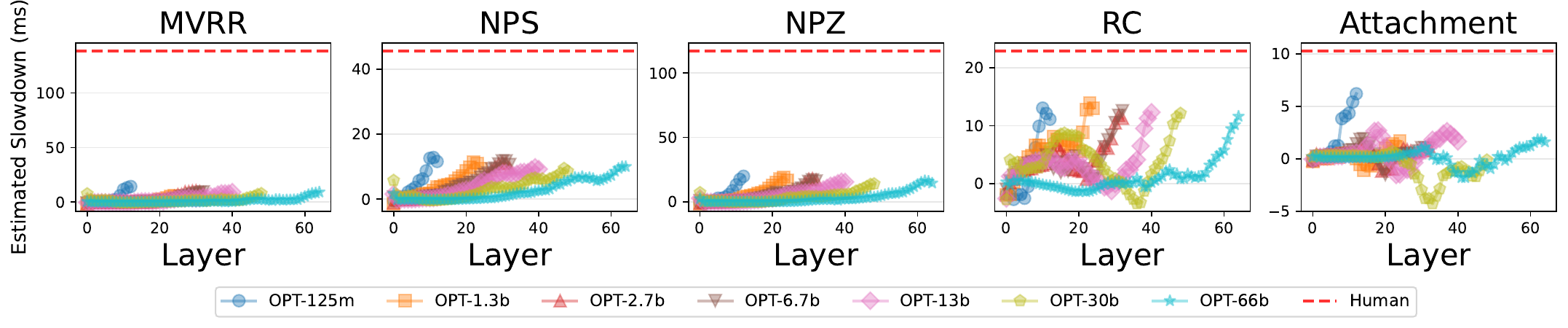}
        \caption{OPT family}
        \label{fig:exp1_opt}
    \end{subfigure}
    
    \caption{Estimated reading time slowdown by layers for each syntactic construction. The red dashed line shows the average observed human slowdown; $y$-axis varies between plots. Later layers show better alignment, but all model families and all layers underestimate the effect.}
    \label{fig:exp1_2}
\end{figure*}

\section{Experiment 1}
\label{sec:exp1}
We first extend existing experiments to compare the reading time slowdown in syntactic ambiguity processing with LM-computed surprisal, focusing on internal layers of LMs, not just the last layer.

\subsection{Settings}
\label{subsec:exp1_settings}

\paragraph{Data}
We use the syntactic ambiguity processing data from~\citet{Huang2024-qe}, which covers five types of syntactically challenging constructions: (i) Main Verb/Reduced Relative (MVRR); (ii) Noun Phrase or Sentential Complement (NPS); (iii) Noun Phrase Complement or Zero Complement (NPZ); (iv) Object/Subject Relative Clause (RC); and (v) High/Low Attachment (Attachment).\footnote{We excluded the Agreement part, which targets the effort for processing grammatical violations, as our initial focus is on syntactic ambiguity processing.}
For each construction, the dataset $D$ contains matched pairs of sentences $(s^+, s^-)_1^{|D|} \in D^+\times D^-$, where $s^+$ is from the syntactically challenging condition and $s^-$ is from an unchallenging condition with explicit cues that resolve syntactic ambiguity (see Table~\ref{tab:examples}). 
For example, in the MVRR construction shown in Table~\ref{tab:examples}, the challenging version ($\in D^+$) is ambiguous: ``fed'' could be either a main transitive verb or a past participle in a relative clause without the relative pronoun and copula. 
This ambiguity is resolved only when readers reach ``remained.''
In contrast, the unambiguous version ($\in D^-$) includes the relative pronoun and copula, making the structure immediately clear. 
Each sentence pair has an annotated disambiguating point $t^*$ where ambiguity is resolved on the $D^+$ side (and its corresponding position on the $D^-$ side), as shown in bold in Table~\ref{tab:examples}.
Slowdowns are observed around $t^*$ in the $D^+$ condition compared to $D^-$, and quantifying this magnitude of slowdown is our focus.

The data contains 24 unique sentence pairs for each syntactic phenomenon, resulting in a total of 120 unique pairs with a total of 3,371 tokens.
Sentences are annotated with token-level human reading times.
Our human reading data comes from \citet{Huang2024-qe}, who used a web-based self-paced reading paradigm on over 2K participants, resulting in around 1.2M data points across all the tokens in the dataset, and around 87K data points at the disambiguating points (including the corresponding point in $D^-$ side).
In our study, as a preprocessing step, reading times are averaged across participants prior to analysis.

\paragraph{Procedure}
Our analysis closely resembles that of~\citet{Wilcox2021-gy}, who estimated predicted reading time slowdowns from LM surprisal.
Let $\bm w=[w_1,\cdots,w_n]^\top$ be tokens in the held-out corpus, and let $\bm y=[y_1,\cdots,y_n]^\top$ be their respective reading times.
For each token, we select a set of word-level linguistic features $\bm f(w_k) \in \mathcal{R}^m$, including its length in characters, unigram frequency, and surprisal. We then fit a regression model to predict by-token reading time $g: \bm f(w_k) \mapsto \hat{y}_k$ from features.\footnote{Following work that establishes a linear link between surprisal and reading times \citep{Shain2022-qv}, we use a linear regression model: $\texttt{RT}(w_t) = \beta_0 + \beta_1 \cdot \texttt{Surprisal}(w_t) + \beta_2 \cdot \texttt{Length}(w_t) + \beta_3 \cdot \texttt{LogFreq}(w_t) + \beta_4 \cdot \texttt{Surprisal}(w_{t-1}) + \beta_5 \cdot \texttt{Length}(w_{t-1}) + \beta_6 \cdot \texttt{LogFreq}(w_{t-1}) + \beta_7 \cdot \texttt{Surprisal}(w_{t-2}) + \beta_8 \cdot \texttt{Length}(w_{t-2}) + \beta_9 \cdot \texttt{LogFreq}(w_{t-2}) + \epsilon.$ (Appendix~\ref{app:regression})}
The regression model is trained on the filler-sentence part of the dataset~\cite{Huang2024-qe}.
Then, this regression model is run on the target data: $s^+=[w_1^+,\cdots,w_n^+]^\top \in D^+$ and $s^-=[w_1^-,\cdots,w_n^-]^\top \in D^-$.  
Estimated reading times for each token are obtained ($\hat{\bm y}^+=[\hat{y}^+_1,\cdots,\hat{y}^+_n]^\top$ and $\hat{\bm y}^-=[\hat{y}^-_1,\cdots,\hat{y}^-_n]^\top$), and we compute the reading time difference at the disambiguating point $t^*$ between the two conditions, yielding surprisal-estimated reading time slowdowns.
Note that we compared the reading time difference summed over $t^*$ and $t^*+1$, given the presence of spillover in the data~\cite{Huang2024-qe}.
We refer to these as regions of interest (RoIs) in this section.
We average this estimated slowdown across all sentence pairs in the dataset.
We repeatedly conducted this procedure using surprisal $S^{(\layer)}_t$ from each layer \layer obtained with logit lens (\cref{subsec:internal_surprisal}) to examine which layer's surprisal better estimates the human reading slowdown between $D^+$ and $D^-$ conditions.

\paragraph{LMs}
We examine 19 open-source Transformer LMs including GPT-2 (124M, 355M, 774M, and 1.5B parameters;~\citealp{Radford_undated-nn}), OPT (125M, 1.3B, 2.7B, 6.7B, 13B, 30B, and 66B parameters;~\citealp{OPT}), Pythia (70M, 160M, 410M, 1B, 1.4B, 2.8B, 6.9B, and 12B parameters;~\citealp{biderman2023Pythia}).
We excluded instruction-tuned models~\cite{kuribayashi-etal-2024-psychometric}.

\paragraph{Surprisal}
We compute layer-specific surprisal $S^{(\layer)}_t$ for each token at each layer using the Logit-Lens method~\cite{logitlens} (\cref{subsec:internal_surprisal}). Due to the known limitations for using Logit-Lens to examine early layers, we conducted an additional analysis using Tuned-Lens~\cite{belrose2023eliciting}, and found no substantial difference in the results (\cref{app:tuned-lens}); thus, results in the main text are all from the simpler, Logit-Lens version.
We applied Whitespace-Trailing Decoding~\cite{Oh2024-cf} to compute the accurate next-word probabilities for exact disambiguating tokens.

\subsection{Results}
Figure~\ref{fig:exp1_2} presents the averaged reading time differences between syntactically challenging ($D^+$) and unchallenging ($D^-$) conditions by LM surprisal across different layers, alongside the actual human reading time difference (red line; note that the $y$-axis differs across phenomena).
First, it is evident that surprisal from all layers underestimates the human reading time difference, consistent with prior findings on targeted misalignment~\cite{Van_Schijndel2021-sm,Huang2024-qe}.
Second, later layers consistently provide relatively better (even if underestimated) predictions of the reading time difference between $D^+$ and $D^-$ conditions.
This contrasts with previous findings on naturalistic reading, where earlier layers typically yielded superior predictions of human reading times~\cite{kuribayashi2025largelanguagemodelshumanlike}.
That is, the best layer is notably later in syntactically challenging contexts compared to naturalistic reading scenarios.

\begin{figure*}[t]
    \centering
    \begin{subfigure}{\textwidth}
        \centering
        \includegraphics[width=\textwidth]{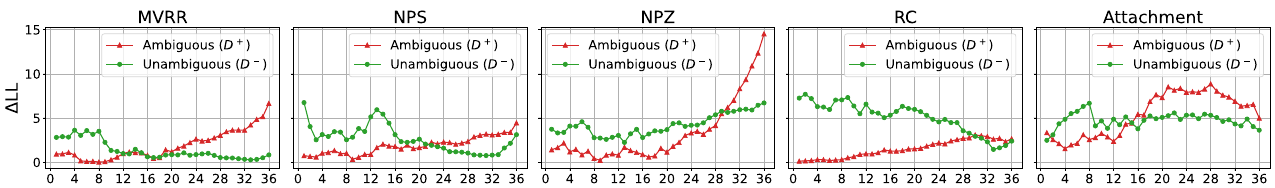}
        \caption{\textsc{RoI} data points}
        \label{fig:ppp_roi}
    \end{subfigure}

    \begin{subfigure}{\textwidth}
        \centering
        \includegraphics[width=\textwidth]{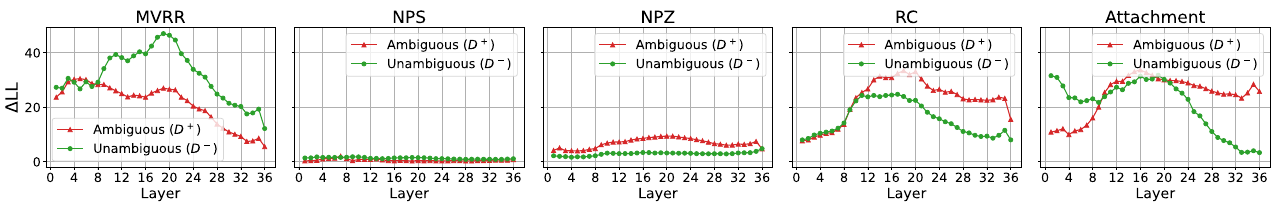}
        \caption{$\overline{\textsc{RoI}}$ data points}
        \label{fig:ppp_out_roi}
    \end{subfigure}

    \caption{By-layer PPP of Pythia 12B in the four conditions: $D^+\cap\textsc{RoI}$ (upper red), $D^-\cap\textsc{RoI}$ (upper green), $D^+\cap\overline{\textsc{RoI}}$ (bottom red), and $D^-\cap\overline{\textsc{RoI}}$ (bottom green). Only in $D^+\cap\textsc{RoI}$, better PPPs are from deeper layers.}
    \label{fig:ppp_graph}
\end{figure*}

\subsection{Interim Discussion}

\paragraph{Syntactic insensitivity of earlier layers.}
Earlier layers' surprisal in RoIs was almost the same between $D^+$ and $D^-$ conditions (Figure~\ref{fig:exp1_2}), leading to a failure in simulating the contrastive reading time slowdown.
One possible explaination is that earlier layers are not sensitive enough to long dependencies and are distracted by local co-occurrences. For example, given the MVRR construction ``The girl fed the lamb \textbf{remained}...,'', an earlier layer might only consider the local co-occurrence of ``the lamb \textbf{remained},'' and therefore assign lower surprisal to ``remained'' even though it is implausible given the larger context.

\paragraph{Dual alignment.}
We also observe a systematic shift in which layer's surprisal best approximates human behavior: later layers are more effective for syntactically challenging constructions, whereas prior work has shown that earlier layers better capture naturalistic reading~\cite{kuribayashi2025largelanguagemodelshumanlike}.
This \textit{dual alignment} suggests that if we attempt to model the reading behavior of syntactically challenging constructions through the lens of prediction, more extensively contextualized representations would be selectively required.
This corresponds to a dual-mechanism perspective on human sentence processing \citep{narayanan1998bayesian, Van_Schijndel2021-sm}, wherein humans may usually read sentences using a relatively shallow processing strategy (aligned with earlier layers) and switch to a deeper, more contextually integrated processing mode (somewhat better aligned with later layers) when confronted with syntactically challenging constructions that require reanalysis or complex integration of contextual information.
The following sections (\cref{sec:exp2} and \cref{sec:exp3}) conduct a follow-up analysis to further explore this hypothesis.

\section{Experiment 2: Psychometric predictive power analysis with layer shift}
\label{sec:exp2}
The results of our previous study challenge the existing paradigm in LM-based cognitive modeling, which uses a single layer to model all data points. Rather, they suggest that the part of the LM used for modeling human cognition may need to change dynamically depending on the phenomena.
Given this view, we conduct a follow-up experiment to extend and clarify these results.
Rather than looking at predicted human reading slowdowns, we instead measure models' psychometric predictive power~\cite[PPP;][]{Goodkind2018PredictiveQuality}, and break results down into our relevant experimental conditions.
Given our earlier results, we expect to find that earlier layers better simulate human reading behavior as observed in naturalistic reading~\cite{kuribayashi2025largelanguagemodelshumanlike}, while deeper processing is recruited for more syntactically challenging sentences; once the difficulty is resolved, earlier layers again become effective.
In this study, we simply ask whether such a shift can be modeled using LMs' layer-wise surprisal, and leave questions about control between shallow vs. deep processing for future research.

\subsection{Problem setting}
\label{subsec:ppp}
For a simple model of context-dependent layer shift, we split the data points into four conditions based on whether they come from ambiguous sentences, and from regions of interest: \{$D^+$, $D^-$\}$\times$\{\textsc{RoI}, $\overline{\textsc{RoI}}$\}.
Then, we examine which layer's surprisal best fits data points in each condition based on PPP.
If our hypothesis --- the advantage of a later layer is a signature of processing difficulty --- holds, we expect that the deeper layer yields a better PPP only for $D^+\times$\textsc{RoI} combination.
Hereafter, we include tokens at $t^*$-2, $t^*$-1, $t^*$, $t^*$+1, and $t^*$+2 as \textsc{RoI} to consider relative reading time magnitude change around the disambiguating point; all other words constitute the $\overline{\textsc{RoI}}$ group.

\begin{table*}[t]
\centering
\scriptsize
\tabcolsep=0.04cm
\begin{tabular}{lrrrrrrrrrrrrrrrrrrrr}
\toprule
 & \multicolumn{4}{c}{MVRR} & \multicolumn{4}{c}{NPS} & \multicolumn{4}{c}{NPZ} & \multicolumn{4}{c}{RC} & \multicolumn{4}{c}{Attachment} \\
\cmidrule(r){2-5} \cmidrule(r){6-9} \cmidrule(r){10-13} \cmidrule(r){14-17} \cmidrule(r){18-21}
 & \multicolumn{2}{c}{$D^+$} & \multicolumn{2}{c}{$D^-$} & \multicolumn{2}{c}{$D^+$} & \multicolumn{2}{c}{$D^-$} & \multicolumn{2}{c}{$D^+$} & \multicolumn{2}{c}{$D^-$} & \multicolumn{2}{c}{$D^+$} & \multicolumn{2}{c}{$D^-$} & \multicolumn{2}{c}{$D^+$} & \multicolumn{2}{c}{$D^-$} \\
 Model & \multicolumn{1}{c}{RoI}  & \multicolumn{1}{c}{$\overline{\text{RoI}}$} & \multicolumn{1}{c}{RoI} & \multicolumn{1}{c}{$\overline{\text{RoI}}$} & \multicolumn{1}{c}{RoI} & \multicolumn{1}{c}{$\overline{\text{RoI}}$} & \multicolumn{1}{c}{RoI} & \multicolumn{1}{c}{$\overline{\text{RoI}}$} & \multicolumn{1}{c}{RoI} & \multicolumn{1}{c}{$\overline{\text{RoI}}$} & \multicolumn{1}{c}{RoI} & \multicolumn{1}{c}{$\overline{\text{RoI}}$} & \multicolumn{1}{c}{RoI} & \multicolumn{1}{c}{$\overline{\text{RoI}}$} & \multicolumn{1}{c}{RoI} & \multicolumn{1}{c}{$\overline{\text{RoI}}$} & \multicolumn{1}{c}{RoI} & \multicolumn{1}{c}{$\overline{\text{RoI}}$} & \multicolumn{1}{c}{RoI} & \multicolumn{1}{c}{$\overline{\text{RoI}}$} \\
\cmidrule(r){1-1} \cmidrule(r){2-5} \cmidrule(r){6-9} \cmidrule(r){10-13} \cmidrule(r){14-17} \cmidrule(r){18-21}
GPT2-sm & 0.46 & $-$0.99 & \textbf{0.61} & $-$0.7 & \textbf{0.87} & $-$0.72 & $-$0.15 & 0.01 & $-$0.69 & 0.05 & \textbf{0.89} & 0.38 & \textbf{0.82} & $-$0.37 & $-$0.87 & $-$0.78 & \textbf{0.72} & 0.39 & $-$0.39 & $-$0.83 \\
GPT2-md & \textbf{0.59} & $-$0.91 & 0.34 & $-$0.87 & \textbf{0.90} & $-$0.68 & $-$0.7 & $-$0.42 & \textbf{0.74} & $-$0.24 & 0.52 & 0.39 & \textbf{0.81} & $-$0.43 & $-$0.84 & $-$0.82 & \textbf{0.42} & $-$0.08 & $-$0.54 & $-$0.95 \\
GPT2-lg & \textbf{0.53} & $-$0.98 & $-$0.33 & $-$0.91 & \textbf{0.88} & $-$0.58 & $-$0.62 & 0.71 & \textbf{0.88} & $-$0.65 & 0.13 & 0.33 & \textbf{0.91} & $-$0.82 & $-$0.85 & $-$0.83 & \textbf{0.08} & $-$0.23 & \textbf{0.08} & $-$0.96 \\
GPT2-xl & \textbf{0.88} & $-$0.96 & $-$0.72 & $-$0.86 & \textbf{$-$0.07} & $-$0.3 & $-$0.85 & $-$0.1 & \textbf{0.88} & $-$0.73 & 0.57 & 0.33 & \textbf{0.96} & $-$0.76 & $-$0.44 & $-$0.85 & \textbf{$-$0.32} & $-$0.45 & $-$0.51 & $-$0.97 \\
\cmidrule(r){2-5} \cmidrule(r){6-9} \cmidrule(r){10-13} \cmidrule(r){14-17} \cmidrule(r){18-21}
OPT-125m & \textbf{0.73} & $-$0.57 & 0.31 & $-$0.26 & \textbf{0.97} & $-$0.14 & $-$0.48 & 0.13 & \textbf{0.89} & $-$0.2 & 0.75 & 0.55 & \textbf{0.81} & $-$0.09 & $-$0.56 & $-$0.12 & 0.51 & \textbf{0.92} & $-$0.91 & $-$0.84 \\
OPT-1.3b & \textbf{0.67} & $-$0.97 & $-$0.77 & $-$0.98 & \textbf{0.81} & 0.52 & 0.23 & $-$0.44 & \textbf{0.83} & $-$0.89 & $-$0.46 & 0.32 & \textbf{0.83} & $-$0.65 & 0.71 & $-$0.75 & \textbf{0.40} & $-$0.0 & $-$0.65 & $-$0.97 \\
OPT-2.7b & \textbf{0.70} & $-$0.95 & $-$0.78 & $-$0.98 & \textbf{0.86} & 0.35 & 0.14 & $-$0.27 & \textbf{0.86} & $-$0.89 & 0.21 & $-$0.3 & \textbf{0.81} & $-$0.15 & 0.57 & $-$0.48 & \textbf{0.18} & 0.05 & $-$0.63 & $-$0.96 \\
OPT-6.7b & \textbf{0.23} & $-$0.96 & 0.06 & $-$0.98 & \textbf{0.74} & 0.43 & $-$0.4 & $-$0.42 & \textbf{0.49} & $-$0.91 & $-$0.57 & $-$0.35 & \textbf{0.87} & $-$0.59 & 0.25 & $-$0.73 & \textbf{$-$0.20} & $-$0.26 & $-$0.26 & $-$0.93 \\
OPT-13b & \textbf{0.09} & $-$0.96 & $-$0.71 & $-$0.97 & \textbf{0.71} & 0.14 & 0.15 & $-$0.76 & \textbf{0.81} & $-$0.87 & $-$0.19 & 0.03 & \textbf{0.88} & $-$0.65 & 0.74 & $-$0.83 & \textbf{0.26} & $-$0.43 & $-$0.22 & $-$0.97 \\
OPT-30b & $-$0.13 & $-$0.86 & \textbf{0.29} & $-$0.9 & \textbf{0.61} & 0.39 & $-$0.52 & 0.16 & 0.13 & $-$0.76 & $-$0.08 & \textbf{0.32} & \textbf{0.82} & $-$0.61 & $-$0.38 & $-$0.8 & $-$0.42 & \textbf{$-$0.22} & $-$0.35 & $-$0.91 \\
OPT-66b & \textbf{0.07} & $-$0.88 & $-$0.07 & $-$0.83 & 0.59 & \textbf{0.65} & $-$0.49 & $-$0.49 & \textbf{0.48} & $-$0.96 & $-$0.2 & $-$0.61 & \textbf{0.77} & $-$0.36 & 0.53 & $-$0.59 & \textbf{0.66} & 0.32 & 0.19 & $-$0.93 \\
\cmidrule(r){2-5} \cmidrule(r){6-9} \cmidrule(r){10-13} \cmidrule(r){14-17} \cmidrule(r){18-21}
PYT-70m & \textbf{0.04} & $-$0.86 & $-$0.02 & $-$0.86 & \textbf{0.74} & $-$0.69 & 0.41 & $-$0.42 & 0.3 & 0.39 & \textbf{0.76} & 0.41 & \textbf{$-$0.13} & $-$0.62 & $-$0.85 & $-$0.87 & \textbf{0.88} & 0.59 & $-$0.51 & $-$0.14 \\
PYT-160m & \textbf{0.31} & $-$0.93 & 0.16 & $-$0.67 & 0.24 & $-$0.11 & \textbf{0.50} & $-$0.49 & 0.29 & 0.19 & 0.09 & \textbf{0.58} & \textbf{0.52} & $-$0.5 & $-$0.76 & $-$0.06 & \textbf{0.91} & 0.6 & $-$0.55 & $-$0.42 \\
PYT-410m & \textbf{0.05} & $-$0.89 & $-$0.34 & $-$0.86 & \textbf{0.82} & $-$0.61 & $-$0.2 & $-$0.55 & 0.65 & 0.58 & 0.65 & \textbf{0.66} & \textbf{0.92} & 0.28 & $-$0.67 & $-$0.33 & 0.06 & \textbf{0.49} & $-$0.69 & $-$0.94 \\
PYT-1b & \textbf{0.70} & $-$0.95 & $-$0.46 & $-$0.94 & \textbf{0.69} & $-$0.01 & $-$0.58 & $-$0.62 & \textbf{0.57} & $-$0.64 & $-$0.56 & 0.17 & \textbf{0.90} & $-$0.87 & $-$0.81 & $-$0.89 & \textbf{0.70} & 0.6 & $-$0.76 & $-$0.85 \\
PYT-1.4b & \textbf{0.55} & $-$0.96 & $-$0.15 & $-$0.91 & \textbf{0.80} & $-$0.27 & 0.51 & $-$0.57 & \textbf{0.50} & $-$0.15 & $-$0.71 & 0.01 & \textbf{0.91} & 0.8 & $-$0.86 & $-$0.39 & \textbf{0.46} & 0.28 & $-$0.83 & $-$0.9 \\
PYT-2.8b & \textbf{0.80} & $-$0.95 & $-$0.46 & $-$0.88 & 0.44 & $-$0.58 & \textbf{0.51} & $-$0.4 & \textbf{0.64} & $-$0.65 & $-$0.54 & 0.25 & \textbf{0.90} & $-$0.68 & $-$0.94 & $-$0.85 & \textbf{0.72} & 0.3 & $-$0.46 & $-$0.86 \\
PYT-6.9b & \textbf{0.54} & $-$0.75 & $-$0.21 & $-$0.82 & \textbf{0.96} & $-$0.56 & $-$0.7 & $-$0.82 & 0.75 & 0.55 & 0.7 & \textbf{0.96} & \textbf{0.94} & 0.28 & $-$0.94 & $-$0.54 & \textbf{0.82} & 0.53 & $-$0.3 & $-$0.77 \\
PYT-12b & \textbf{0.88} & $-$0.89 & $-$0.83 & $-$0.41 & \textbf{0.93} & $-$0.43 & $-$0.69 & $-$0.82 & \textbf{0.79} & 0.38 & 0.78 & 0.74 & \textbf{0.97} & 0.46 & $-$0.92 & $-$0.22 & \textbf{0.80} & 0.56 & 0.04 & $-$0.76 \\
\bottomrule
\end{tabular}
\caption{Correlation between layer depth and PPP by model and condition. $D^+\cap\textsc{RoI}$ exhibits positive correlations.}
\label{tab:correlation}
\end{table*}

\paragraph{Psychometric predictive power (PPP)}
We quantify the goodness-of-fit of layer-specific surprisal to data points in each condition: \{$D^+$, $D^-$\}$\times$\{\textsc{RoI}, $\overline{\textsc{RoI}}$\}.
This is measured by a log-likelihood-based score, $\Delta$LL (i.e., psychometric predictive power; PPP), following existing studies~\cite{Goodkind2018PredictiveQuality,Wilcox2020OnBehavior,kuribayashi-etal-2021-lower,kuribayashi-etal-2022-context,kuribayashi2025largelanguagemodelshumanlike,Oh2023-zw}.\footnote{The total $\Delta$LL over the dataset, not token-level average.}
Specifically, we fit two linear regression models to predict word-by-word reading times: a full model that includes both surprisal and baseline linguistic features, and a reduced model that includes only baseline features.\footnote{We use the same regression model as in \cref{sec:exp1}, and for baseline features, we excluded all the surprisal factors.}
The PPP score is defined as the difference in log-likelihood between these two models: $\Delta \text{LL} = \text{LL}_{\text{full}} - \text{LL}_{\text{baseline}}$, which quantifies how much the addition of surprisal improves the model fit (see Appendix A in~\citealp{kuribayashi2025largelanguagemodelshumanlike}).
Higher $\Delta$LL values indicate that surprisal better captures human reading behavior.
We repeatedly compute PPP for each layer in each of the four conditions: \{$D^+$, $D^-$\}$\times$\{\textsc{RoI}, $\overline{\textsc{RoI}}$\}.

\paragraph{Measure}
In each condition, we report Pearson's correlation coefficient between layer depth and PPP.
A higher correlation indicates the tendency of later layers to better simulate the respective reading time, which is expected only in the $D^+ \cap$ \textsc{RoI} condition.

\subsection{Results}
\label{subsec:exp2_results}
Let us begin with observing a representative pattern from Pythia 12B model in the four different conditions (Figure~\ref{fig:ppp_graph}).
The figure reveals that the increasing PPP toward deeper layers is distinctive in the $D^+\cap\textsc{RoI}$ condition (top red lines).
Table~\ref{tab:correlation} summarizes the correlation between layer depth and $\Delta\mathrm{LL}$, providing a complementary view of the layer-wise trend.
A positive correlation indicates that deeper layers progressively improve the prediction of human reading behavior.
Positive correlations are consistently observed for $D^+\cap\textsc{RoI}$ condition across all models and constructions.
This corroborates that human reading behavior under syntactic ambiguity particularly aligns with deeper layers, consistent with our expectation.

Notably, the contrast in correlation across conditions \{$D^+$, $D^-$\}$\times$\{RoI, $\overline{\text{RoI}}$\} becomes more pronounced in larger models.
For example, Pythia-12B shows correlations of 0.88 (RoI in $D^+$) vs. $-$0.83 (RoI in $D^-$) for MVR, and 0.93 (RoI in $D^+$) vs. $-$0.69 (RoI in $D^-$) for NPS, while these are somewhat attenuated in smaller ones, indicating that as models scale up, they develop a clearer differentiation in how layer depth relates to syntactically challenging versus unchallenging reading.

\section{Experiment 3: Probability-update as processing effort}
\label{sec:exp3}
Our experiments so far have converged on the finding that surprisal from deeper layers with extensive contextualization better captures syntactic ambiguity processing, a behavior that requires substantial cognitive effort.
One interpretation of this result is that it arises because human processing in such regions involves an initial shallow prediction with surface-level features, such as unigram frequency or local co-occurrence information, which must be subsequently revised by considering a broader linguistic context, incurring a higher processing cost.
In this section, we propose a method for identifying such data points by inspecting the difference in predictions between LM layers, i.e., the advantage that deeper contextualization buys you for prediction.

\begin{figure*}[t]
    \centering
    \includegraphics[width=0.99\textwidth]{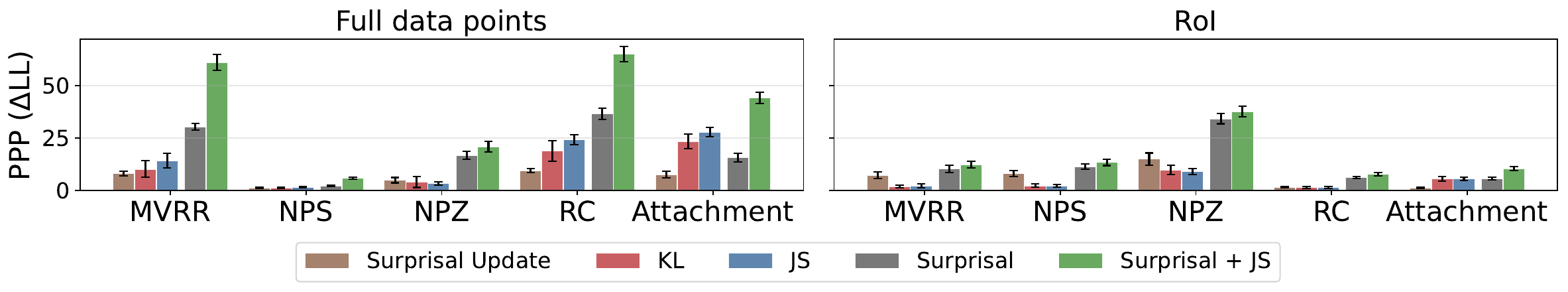}
    \caption{PPP obtained by probability-update measurements introduced in~\cref{subsec:probability_update} on ten different data conditions \{5 phenomena\}$\times$\{Full, RoI\}. PPP scores from 19 LMs are averaged, and error bars indicate 95\% confidence interval. PPP with surprisal (gray) and the one with both surprisal and JS divergence (green) are also reported.}
    \label{fig:fig_kl}
\end{figure*}

\subsection{Measurements}
\label{subsec:probability_update}
We explore several information-theoretic measures that quantify the change in predictive distributions between shallow and deep processing.
Our first formulation computes the \emph{change} in surprisal of a word \word between a shallow and a deep layer, rather than just computing surprisal at a certain layer.
We term this the \defn{surprisal update} ($\mathrm{SU}$) defined as:

\begin{align}
    \nonumber
    &\mathrm{SU}(\word|\prevwords) = S_t^{\mathrm{\textcolor{\orange}{shallow}}} - S_t^{\mathrm{\textcolor{\orange}{deep}}} \\ 
    \nonumber
    &= -\log P_t(\word|\prevwords) - \left(-\log Q_t(\word|\prevwords)\right) \\ 
    &= \log\frac{Q_t(\word|\prevwords)}{P_t(\word|\prevwords)} \;\;\mathrm{.}
\end{align}

\noindent
where $P_t(\word|\prevwords)$ and $Q_t(\word|\prevwords)$ denote the probability of $\word$ in context estimated by the \textcolor{\orange}{first} and \textcolor{\orange}{last} layer, respectively.\footnote{In the regression analysis, we applied Z-score normalization to each layer's surprisals before computing surprisal-update to mitigate layer-dependent scale differences.}
We henceforth denote the corresponding probability distributions over the model's subword vocabulary as $P_t$ and $Q_t$.

As a second measure, we extend surprisal update to the full next-(sub)word distribution using the \textbf{Kullback-Leibler divergence} (KL), i.e., the expected value of SU under $Q_t$:
\begin{align}
    \nonumber
    \mathrm{KL}(Q_t || P_t) &= \sum_{\w \in \Word}Q_t(\w|\prevwords) \log\frac{Q_t(\w|\prevwords)}{P_t(\w|\prevwords)}  \\
    &=\mathbb{E}_{\w \sim Q_t } \left[ \mathrm{SU}(\w|\prevwords)\right] 
    \;\;\mathrm{.}
\end{align}

\noindent
This quantifies the discrepancy between the final prediction $Q_t$ and the initial one $P_t$.
As the KL is asymmetric, we also examine its symmetric version, namely, \textbf{Jensen--Shannon divergence} (JS):
\begin{align}
    \nonumber
    \mathrm{JS}(Q_t || P_t) &= \frac{1}{2} \left( \mathrm{KL}(Q_t || M_t) + \mathrm{KL}(P_t || M_t)  \right) \\
    M_t &:= \frac{1}{2} \left( P_t + Q_t \right) \;\;\mathrm{.}
\end{align}

We compute these measures for each token\footnote{KL and JS are computed for each subword position, and if a token consists of multiple subwords, we simply sum up the subword-level scores, similarly to cumulative surprisal.} and assess their effectiveness in modeling human reading times.
We hypothesize that larger discrepancies between shallow and deep predictions, as quantified by SU, KL, and JS, may be associated with higher processing effort, as they indicate a greater room to revise initial predictions based on broader context.
Conversely, if surprisal or probability distribution does not change much after extensive contextualization, it suggests that the word is easier to integrate, requiring less cognitive effort.
Our proposal is similar in spirit to the information-theoretic model of shallow vs. deep processing presented in \citet{li2024information}.

\subsection{Problem setting}
As in Experiment 2, we model reading time in data from~\citet{Huang2024-qe} using a linear regression model.
We replace surprisal with one of our probability-update predictors (i.e., SU, KL, or JS).
To evaluate whether the probability-update captures processing effort difference across challenging and unchallenging conditions, we report results for RoI regions (as defined in \cref{sec:exp2}) as well as for full data points, yielding ten conditions: \{5 phenomena\}$\times$\{Full, RoI\}.
Note that each condition includes both syntactically challenging and unchallenging items.

\subsection{Results}
Figure~\ref{fig:fig_kl} shows the average PPP across 19 models for each data condition.
Detailed results with likelihood ratio test~\cite{Wilks1938-wt} are also shown in Tables~\ref{tab:ppp_suprisal_update},~\ref{tab:ppp_kl}, and~\ref{tab:ppp_js} in Appendix~\ref{app:ppp_prob_update}.
Consistent with our hypothesis, probability-update measures significantly improve the model fit in most cases, in both full and RoI conditions.
This indicates that words associated with larger probability updates correspond to increased reading times in humans.
Among the three measures, JS typically exhibited the best PPP empirically.

Nevertheless, following the setting of~\cref{sec:exp1}, the estimated slowdown by probability-update measures was under 10 ms (Appendix~\ref{app:effect_prob_update}).
In addition, as shown in Figure~\ref{fig:fig_kl}, PPP gains are more nuanced for the RoI data points compared to the full data conditions.
Thus, we tentatively conclude that, while probability-update measures explored in this section are potentially effective features for reading time modeling, their advantage is not specifically associated with the processing cost of syntactically ambiguous sentences.

\subsection{Analysis}
\label{subsec:exp3:analysis}
Is the advantage of the probability-update measure in terms of PPP orthogonal to surprisal?
Taking the JS value as an example, we evaluate the additive effect of the JS to the last layer's surprisal.
Figure~\ref{fig:fig_kl} also includes the results for surprisal (gray) and both surprisal and JS (green), where these features are added to the same baseline regression model (\cref{subsec:exp1_settings}).
As shown in Figure~\ref{fig:fig_kl}, the Surprisal + JS (green) setting typically exhibits an advantage over the Surprisal ones, supporting the complementary effect of JS-based probability-update to the commonly-used surprisal feature.
In Appendix~\ref{app:ppp_prob_update_additive}, likelihood ratio tests are conducted between the two nested regression models: one with surprisal plus baseline features and the other with surprisal, JS, and baseline features, showing that MVR (Full), RC (Full), and Attachment (Full/RoI) settings tend to yield statistical significance.

\section{Discussion}
\label{sec:discussion}
We have investigated the alignment between internal-layer surprisal from LMs and human reading behavior in syntactically challenging constructions.
It is a natural extension of previous work connecting layer-wise dynamics to \emph{offline} language processing~\cite{Tenney2019-bb,he-etal-2024-decoding,hu2025signatures}.
On the one hand, the dual alignment we observe is consistent with the shallow vs. deep processing that characterizes human language comprehension \citep{Barton1993-rz,Christianson2001-sk,ferreira+2002good}.
This suggests that LMs have stages of processing, perhaps analogous to the two-stage model of human sentence processing (see discussion in \citealp{Van_Schijndel2021-sm}).
On the other hand, our results pose challenges to cognitive modeling with LMs.
First, while human processing dynamics unfold over time, model dynamics unfold over internal layers, thus requiring another (perhaps thorny) theoretical link between the two.
Second, we are unable to identify a single component of an LM that is the optimal basis for modeling (all) human language processing behavior.
While early layers function best for naturalistic reading are relatively worse predictors for reading times of ambiguous sentences.
Finally, in human reading, reanalysis is thought to be a specialized operation, triggered by abnormally high surprisal \cite{Warner1987-wv,Levy2008-pl} or entropy \cite{botvinick2001-lb,Ness2025-pu}.
In contrast, LM architectures apply essentially the same computational operations, i.e., highly contextualized processing, to all inputs.

Our contribution sharpens what commitments must be made about LMs when using them as cognitive models of human processing.
Rather than using LM surprisal as one single model to capture human sentence processing, we should look to \emph{parts} of LMs as models of sub-processes or components of human language processing. 
In this spirit, we suggest several probability-update measures over internal layers as a way to model the cognitive \textit{distance} between early- and late-stage processing when processing a word.

Our exploration of probability updates across layers in relation to cognitive cost (\cref{sec:exp3}) connects to the Bayesian approaches to sentence processing~\cite{Ratcliff1978-dr,narayanan1998bayesian,Levy2008-xa,Ratcliff2008-dn,norris2006bayesian,Norris2009-pt,Itti2009-rk}.
While these theories typically focus on incremental, input-driven updates to the probability distribution, our analysis may align with the memory-based contextual integration, whereby internal layers iteratively refine representations and update predictive distribution.
Establishing theoretical links between our approach and psycholinguistic theories is an important future work, both for better situating LMs as a tool for psycholinguistics and for addressing an NLP interpretability question --- if the goal is to identify syntactic ambiguity processing within an LM, which approach is more appropriate?

\section{Conclusions}
\label{sec:conclusion}
This study provides evidence that surprisal from later layers with richer contextualization better captures human ambiguity processing.
This contrasts with prior work on holistic reading-time modeling, which reported stronger alignment with earlier-layer surprisal~\cite{kuribayashi2025largelanguagemodelshumanlike}, and we introduce a \textit{dual alignment} between LM layers and human sentence processing.
These findings suggest that, rather than treating LM surprisal as a single unified predictor of human sentence processing, it may be more fruitful to look to \emph{parts} of LMs as models of sub-processes or components of human language processing.
In this spirit, we propose probability-update measures across layers as a way to quantify the cognitive distance between early- and late-stage processing, demonstrating the potential of layer-contrastive information-theoretic measures for modeling sentence processing effort.

\section*{Limitations}
We only investigated English LMs and human reading data in the English syntactic ambiguity processing.
Extending the analysis to other languages with different syntactic structures and ambiguity types will enhance the universality of our findings.
We used self-paced reading time data from~\citet{Huang2024-qe}, and a concurrent study has released eye-tracking corpora~\cite{Timkey2025-cn}.
The use of eyetracking data will provide fine-grained information about layer-wise alignment; for example, it is likely that first-pass forward reading time aligns with earlier layers, while later regressive behavior relatively aligns with deeper layers.
Integrating reading behavior data over grammatical violations~\cite{Wilcox2021-gy}, which has also been underestimated by surprisal, will be worth exploring, and similar results may be expected, given that earlier layers' surprisal was not sensitive to grammatical structure. 
More generally, our paper alone does not answer whether the dual alignment is specifically related to syntactic ambiguity or generalizes to other well-studied sources of processing difficulty, such as agreement attraction, long-distance dependencies, or similarity-based interference.
More controlled experiments will clarify our findings.

On the LM side, there are several technical issues.
First, internal probabilities are obtained from logit-lens~\cite{logitlens}.
Although, at least compared to tuned-lens~\cite{belrose2023eliciting}, the selection of the probability extraction method did not substantially affect the results, our analysis relies heavily on the method for obtaining internal probabilities, potentially leading to methodological biases. 
Second, for KL- and JS-based probability update measures, we treat the LMs' subword vocabulary as the vocabulary space, which would not be cognitively plausible.
More generally, the token granularity of LMs has been reported to affect the quality of information-theoretic values, e.g., surprisal, and cognitive modeling results~\cite{nair2023words,oh-schuler-2025-impact}, and this issue also applies to our experiments.
Third, exploring other variants of layer-contrastive information-theoretic measures beyond the metrics studied in this paper and establishing their theoretical connections to psycholinguistic theories will also be a promising future direction.

\section*{Ethical Statement}
This study conducts the analysis of publicly available datasets of human reading behavior and LMs.
We expect that these artifacts were collected and released following appropriate ethical guidelines in existing studies.

\section*{AI Writing/Coding Assistance Policy}
We used generative AI tools solely for the purpose of adjusting the grammar and phrasing of the manuscript, and a coding assistant for formatting tables and figures.

\section*{Acknowledgements}
This work was supported by JSPS Grant-in-Aid for Early Career Scientists Grant Number JP23K16938, JSPS KAKENHI Grant Number JP24H00087, JST CREST Grant Number JPMJCR2565, JST PRESTO Grant Number JPMJPR21C2, and JST BOOST Grant Number JPMJBY24B2.

\bibliography{custom}

@article{frank2011insensitivity,
    title = {Insensitivity of the Human Sentence-Processing System to Hierarchical Structure},
    year = {2011},
    journal = {Psychological Science},
    author = {Frank, Stefan L and Bod, Rens},
    number = {6},
    pages = {829--834},
    volume = {22},
    publisher = {Sage Publications Sage CA: Los Angeles, CA},
    url = {https://www.researchgate.net/publication/51140976_Insensitivity_of_the_Human_Sentence-Processing_System_to_Hierarchical_Structure}
}

@inproceedings{kennedy2003dundee,
    title = {{The dundee corpus}},
    year = {2003},
    booktitle = {Proceedings of the 12th European conference on eye movement},
    author = {Kennedy, Alan and Hill, Robin and Pynte, Joël}
}

@article{Futrell2020DependencyOrder,
    title = {{Dependency locality as an explanatory principle for word order}},
    year = {2020},
    journal = {Journal of Language},
    author = {Futrell, Richard and Levy, Roger P. and Gibson, Edward},
    doi = {10.1353/lan.2020.0024},
    issn = {15350665}
}

@article{Levy2008Expectation-basedComprehension,
    title = {{Expectation-based syntactic comprehension}},
    year = {2008},
    journal = {Journal of Cognition},
    author = {Levy, Roger},
    number = {3},
    pages = {1126--1177},
    volume = {106},
    doi = {10.1016/j.cognition.2007.05.006},
    issn = {00100277},
    pmid = {17662975}
}

@inproceedings{Hale2018FindingSearch,
    title = {{Finding Syntax in Human Encephalography with Beam Search}},
    year = {2018},
    booktitle = {Proceedings of ACL 2018},
    author = {Hale, John and Dyer, Chris and Kuncoro, Adhiguna and Brennan, Jonathan R.},
    pages = {2727--2736},
    isbn = {9781948087322},
    doi = {10.18653/v1/p18-1254},
    arxivId = {1806.04127}
}

@MISC{Radford_undated-nn,
  title={Language models are unsupervised multitask learners},
  author={Radford, Alec and Wu, Jeffrey and Child, Rewon and Luan, David and Amodei, Dario and Sutskever, Ilya and others},
  howpublished={OpenAI blog},
  volume={1},
  number={8},
  pages={9},
  year={2019},
  url={https://d4mucfpksywv.cloudfront.net/better-language-models/language_models_are_unsupervised_multitask_learners.pdf}
}

@inproceedings{Wilcox2020OnBehavior,
    title = {{On the Predictive Power of Neural Language Models for Human Real-Time Comprehension Behavior}},
    year = {2020},
    booktitle = {Proceedings of CogSci 2020},
    author = {Wilcox, Ethan Gotlieb and Gauthier, Jon and Hu, Jennifer and Qian, Peng and Levy, Roger},
    pages = {1707--1713},
    url = {http://arxiv.org/abs/2006.01912},
    arxivId = {2006.01912}
}

@inproceedings{Goodkind2018PredictiveQuality,
    title = {{Predictive power of word surprisal for reading times is a linear function of language model quality}},
    year = {2018},
    booktitle = {Proceedings of CMCL},
    author = {Goodkind, Adam and Bicknell, Klinton},
    pages = {10--18},
    doi = {10.18653/v1/w18-0102}
}

@inproceedings{wolf2019transformers,
    title = "Transformers: State-of-the-Art Natural Language Processing",
    author = "Wolf, Thomas  and
      Debut, Lysandre  and
      Sanh, Victor  and
      Chaumond, Julien  and
      Delangue, Clement  and
      Moi, Anthony  and
      Cistac, Pierric  and
      Rault, Tim  and
      Louf, Remi  and
      Funtowicz, Morgan  and
      Davison, Joe  and
      Shleifer, Sam  and
      von Platen, Patrick  and
      Ma, Clara  and
      Jernite, Yacine  and
      Plu, Julien  and
      Xu, Canwen  and
      Le Scao, Teven  and
      Gugger, Sylvain  and
      Drame, Mariama  and
      Lhoest, Quentin  and
      Rush, Alexander",
    editor = "Liu, Qun  and
      Schlangen, David",
    booktitle = "Proceedings of EMNLP 2020: System Demonstrations",
    month = oct,
    year = "2020",
    url = "https://aclanthology.org/2020.emnlp-demos.6/",
    doi = "10.18653/v1/2020.emnlp-demos.6",
    pages = "38--45"
}

@article{Crocker2010-cp,
  title={Computational Psycholinguistics},
  author={Crocker, Matthew W},
  journal={The Handbook of Computational Linguistics and Natural Language Processing},
  year={2007},
  url={https://www.coli.uni-saarland.de/~crocker/documents/crocker-nlp-handbook.pdf}
}

@article{ferreira+2002good,
author = {Fernanda Ferreira and Karl G.D. Bailey and Vittoria Ferraro},
title ={Good-Enough Representations in Language Comprehension},

journal = {Current Directions in Psychological Science},
volume = {11},
number = {1},
pages = {11-15},
year = {2002},
doi = {10.1111/1467-8721.00158},

URL = { 
    
        https://doi.org/10.1111/1467-8721.00158
    
    

},
eprint = { 
    
        https://doi.org/10.1111/1467-8721.00158
    
    

}
}

@inproceedings{li2024information,
  title={An information-theoretic model of shallow and deep language comprehension},
  author={Li, Jiaxuan and Futrell, Richard},
  booktitle={Proceedings of the Annual Meeting of the Cognitive Science Society},
  volume={46},
  year={2024},
  url={https://escholarship.org/content/qt1fd682nd/qt1fd682nd_noSplash_e733707513b8f4be3d407d3f029acd2b.pdf?t=sgric3}
}

@INPROCEEDINGS{Wilcox2021-gy,
  title     = "A Targeted Assessment of Incremental Processing in Neural
               Language Models and Humans",
  booktitle = "Proceedings of ACL 2021",
  author    = "Wilcox, Ethan and Vani, Pranali and Levy, Roger",
  pages     = "939--952",
  month     =  aug,
  year      =  2021,
  url = "https://aclanthology.org/2021.acl-long.76/"
}

@ARTICLE{Clark2013-xs,
  title    = "Whatever next? Predictive brains, situated agents, and the future
              of cognitive science",
  author   = "Clark, Andy",
  journal  = "Behav. Brain Sci.",
  volume   =  36,
  number   =  3,
  pages    = "181--204",
  month    =  jun,
  year     =  2013,
  language = "en",
  url="https://www.cambridge.org/core/journals/behavioral-and-brain-sciences/article/whatever-next-predictive-brains-situated-agents-and-the-future-of-cognitive-science/33542C736E17E3D1D44E8D03BE5F4CD9"
}

@ARTICLE{Van_Schijndel2021-sm,
  title    = "{Single-Stage} Prediction Models Do Not Explain the Magnitude of
              Syntactic Disambiguation Difficulty",
  author   = "van Schijndel, Marten and Linzen, Tal",
  journal  = "Cognitive Science",
  volume   =  45,
  number   =  6,
  pages    = "e12988",
  month    =  jun,
  year     =  2021,
  language = "en",
  url="https://pubmed.ncbi.nlm.nih.gov/34170031/"
}

@inproceedings{kuribayashi-etal-2021-lower,
    title = "Lower Perplexity is Not Always Human-Like",
    author = "Kuribayashi, Tatsuki  and
      Oseki, Yohei  and
      Ito, Takumi  and
      Yoshida, Ryo  and
      Asahara, Masayuki  and
      Inui, Kentaro",
    booktitle = "Proceedings of ACL-IJCNLP 2021",
    month = aug,
    year = "2021",
    url = "https://aclanthology.org/2021.acl-long.405",
    doi = "10.18653/v1/2021.acl-long.405",
    pages = "5203--5217",
}

@inproceedings{futrell-etal-2018-natural,
    title = "The {Natural Stories Corpus}",
    author = "Futrell, Richard  and
      Gibson, Edward  and
      Tily, Harry J.  and
      Blank, Idan  and
      Vishnevetsky, Anastasia  and
      Piantadosi, Steven  and
      Fedorenko, Evelina",
    booktitle = "Proceedings of LREC 2018",
    month = may,
    year = "2018",
    pages = "76--82",
    url = "https://aclanthology.org/L18-1012",
}

@inproceedings{kuribayashi-etal-2022-context,
    title = "Context Limitations Make Neural Language Models More Human-Like",
    author = "Kuribayashi, Tatsuki  and
      Oseki, Yohei  and
      Brassard, Ana  and
      Inui, Kentaro",
    booktitle = "Proceedings of EMNLP 2022",
    month = dec,
    year = "2022",
    url = "https://aclanthology.org/2022.emnlp-main.712",
    doi = "10.18653/v1/2022.emnlp-main.712",
    pages = "10421--10436",
}

@article{opt,
  author       = {Susan Zhang and
                  Stephen Roller and
                  Naman Goyal and
                  Mikel Artetxe and
                  Moya Chen and
                  Shuohui Chen and
                  Christopher Dewan and
                  Mona T. Diab and
                  Xian Li and
                  Xi Victoria Lin and
                  Todor Mihaylov and
                  Myle Ott and
                  Sam Shleifer and
                  Kurt Shuster and
                  Daniel Simig and
                  Punit Singh Koura and
                  Anjali Sridhar and
                  Tianlu Wang and
                  Luke Zettlemoyer},
  title        = {{OPT: Open Pre-trained Transformer Language Models}},
  year         = {2022},
  url          = {https://arxiv.org/abs/2205.01068v4},
  journal = {arXiv preprint},
  volume = {cs.CL/2205.01068v4},
}

@article{Shain2022-qv,
  title={Large-scale evidence for logarithmic effects of word predictability on reading time},
  author={Shain, Cory and Meister, Clara and Pimentel, Tiago and Cotterell, Ryan and Levy, Roger},
  journal={Proceedings of the National Academy of Sciences},
  volume={121},
  number={10},
  pages={e2307876121},
  year={2024},
  publisher={National Acad Sciences},
  url={https://www.pnas.org/doi/abs/10.1073/pnas.2307876121}
}

@inproceedings{de-varda-marelli-2023-scaling,
    title = "Scaling in Cognitive Modelling: a Multilingual Approach to Human Reading Times",
    author = "de Varda, Andrea  and
      Marelli, Marco",
    booktitle = "Proceedings of ACL 2023",
    month = jul,
    year = "2023",
    url = "https://aclanthology.org/2023.acl-short.14",
    doi = "10.18653/v1/2023.acl-short.14",
    pages = "139--149",
}

@ARTICLE{Oh2023-zw,
  title     = "Why does surprisal from larger Transformer-based language models
               provide a poorer fit to human reading times?",
  author    = "Oh, Byung-Doh and Schuler, William",
  journal   = "TACL",
  publisher = "MIT Press",
  volume    =  11,
  pages     = "336--350",
  month     =  mar,
  year      =  2023,
  copyright = "https://creativecommons.org/licenses/by/4.0/",
  language  = "en",
  url = "https://direct.mit.edu/tacl/article/doi/10.1162/tacl_a_00548/115371/Why-Does-Surprisal-From-Larger-Transformer-Based"
}

@ARTICLE{Demberg2008-fd,
  title    = "Data from eye-tracking corpora as evidence for theories of
              syntactic processing complexity",
  author   = "Demberg, Vera and Keller, Frank",
  journal  = "Cognition",
  volume   =  109,
  number   =  2,
  pages    = "193--210",
  month    =  nov,
  year     =  2008,
  language = "en",
  url = {https://www.sciencedirect.com/science/article/abs/pii/S0010027708001741}
}

@article{Wilcox2023-pi,
    author = {Wilcox, Ethan G. and Pimentel, Tiago and Meister, Clara and Cotterell, Ryan and Levy, Roger P.},
    title = "{Testing the Predictions of Surprisal Theory in 11 Languages}",
    journal = {TACL},
    volume = {11},
    pages = {1451-1470},
    year = {2023},
    month = {12},
    issn = {2307-387X},
    doi = {10.1162/tacl_a_00612},
    url = {https://doi.org/10.1162/tacl\_a\_00612},
    eprint = {https://direct.mit.edu/tacl/article-pdf/doi/10.1162/tacl\_a\_00612/2196877/tacl\_a\_00612.pdf},
}

@inproceedings{nair2023words,
  title={Words, Subwords, and Morphemes: What Really Matters in the Surprisal-Reading Time Relationship?},
  author={Nair, Sathvik and Resnik, Philip},
  booktitle={Findings of EMNLP2023},
  year={2023},
  url={https://arxiv.org/abs/2310.17774}
}

@misc{robyn_speer_2022_7199437,
  author       = {Robyn Speer},
  title        = {rspeer/wordfreq: v3.0},
  month        = sep,
  year         = 2022,
  publisher    = {Zenodo},
  version      = {v3.0.2},
  doi          = {10.5281/zenodo.7199437},
  url          = {https://doi.org/10.5281/zenodo.7199437}
}

@inproceedings{seabold2010statsmodels,
  title={statsmodels: Econometric and statistical modeling with {Python}},
  author={Seabold, Skipper and Perktold, Josef},
  booktitle={9th Python in Science Conference},
  year={2010},
url={https://www.statsmodels.org/stable/index.html}
}

@misc{spacy,
title="{spaCy}: Industrial-strength Natural Language Processing in Python",
author = "Honnibal, Matthew and Montani, Ines and Van Landeghem, Sofie and Boyd, Adriane",
year = 2020,
}

@inproceedings{kuribayashi-etal-2024-psychometric,
    title = "Psychometric Predictive Power of Large Language Models",
    author = "Kuribayashi, Tatsuki  and
      Oseki, Yohei  and
      Baldwin, Timothy",
    editor = "Duh, Kevin  and
      Gomez, Helena  and
      Bethard, Steven",
    booktitle = "Findings of NAACL 2024",
    month = jun,
    year = "2024",
    url = "https://aclanthology.org/2024.findings-naacl.129",
    doi = "10.18653/v1/2024.findings-naacl.129",
    pages = "1983--2005",
}

@inproceedings{narayanan1998bayesian,
  title={Bayesian models of human sentence processing},
  author={Narayanan, Srini and Jurafsky, Daniel},
  booktitle={Proceedings of CogSci 1998},
  pages={752--757},
  year={1998},
  organization={Routledge},
  url={https://escholarship.org/uc/item/559732tp#main}
}

@inproceedings{langedijk-etal-2024-decoderlens,
    title = "{D}ecoder{L}ens: Layerwise Interpretation of Encoder-Decoder Transformers",
    author = "Langedijk, Anna  and
      Mohebbi, Hosein  and
      Sarti, Gabriele  and
      Zuidema, Willem  and
      Jumelet, Jaap",
    editor = "Duh, Kevin  and
      Gomez, Helena  and
      Bethard, Steven",
    booktitle = "Findings of NAACL 2024",
    month = jun,
    year = "2024",
    url = "https://aclanthology.org/2024.findings-naacl.296/",
    doi = "10.18653/v1/2024.findings-naacl.296",
    pages = "4764--4780"
}

@misc{logitlens,
title= "interpreting {GPT}: {t}he logit lens.",
author="nostalgebraist",
year=2020,
howpublished="Blog post, retrieved 20 June, 2025",
url="https://www.lesswrong.com/posts/AcKRB8wDpdaN6v6ru/interpreting-gpt-the-logit-lens"
}

@inproceedings{biderman2023pythia,
  title={Pythia: A suite for analyzing large language models across training and scaling},
  author={Biderman, Stella and Schoelkopf, Hailey and Anthony, Quentin Gregory and Bradley, Herbie and O’Brien, Kyle and Hallahan, Eric and Khan, Mohammad Aflah and Purohit, Shivanshu and Prashanth, USVSN Sai and Raff, Edward and others},
  booktitle={Proceedings of ICML 2023},
  pages={2397--2430},
  year={2023},
  organization={PMLR},
  url={https://arxiv.org/abs/2304.01373}
}

@ARTICLE{Smith2013-ap,
  title     = "The effect of word predictability on reading time is logarithmic",
  author    = "Smith, Nathaniel J and Levy, Roger",
  journal   = "Cognition",
  publisher = "Elsevier BV",
  volume    =  128,
  number    =  3,
  pages     = "302--319",
  month     =  sep,
  year      =  2013,
  language  = "en",
  url = "https://www.sciencedirect.com/science/article/pii/S0010027713000413"
}

@article{belrose2023eliciting,
  title={Eliciting Latent Predictions from Transformers with the Tuned Lens},
  author={Belrose, Nora and Furman, Zach and Smith, Logan and Halawi, Danny and McKinney, Lev and Ostrovsky, Igor and Biderman, Stella and Steinhardt, Jacob},
  year={2023},
  journal={arXiv preprint.},
  URL={https://arxiv.org/abs/2303.08112}
}

@INPROCEEDINGS{Tenney2019-bb,
  title     = "{BERT} rediscovers the classical {NLP} pipeline",
  author    = "Tenney, Ian and Das, Dipanjan and Pavlick, Ellie",
  booktitle = "Proceedings of ACL 2019",
  pages     = "4593--4601",
  year      =  2019,
  url = "https://aclanthology.org/P19-1452/"
}

@ARTICLE{Huang2024-qe,
  title    = "Large-scale benchmark yields no evidence that language model
              surprisal explains syntactic disambiguation difficulty",
  author   = "Huang, Kuan-Jung and Arehalli, Suhas and Kugemoto, Mari and
              Muxica, Christian and Prasad, Grusha and Dillon, Brian and Linzen,
              Tal",
  journal  = "Journal of Memory and Language",
  volume   =  137,
  pages    =  104510,
  month    =  aug,
  year     =  2024,
  url = "https://www.sciencedirect.com/science/article/abs/pii/S0749596X24000135"
}

@INPROCEEDINGS{Oh2024-cf,
  title     = "Leading Whitespaces of Language Models’ Subword Vocabulary Pose a
               Confound for Calculating Word Probabilities",
  author    = "Oh, Byung-Doh and Schuler, William",
  booktitle = "Proceedings of EMNLP 2024",
  pages     = "3464--3472",
  month     =  nov,
  year      =  2024,
  url = "https://aclanthology.org/2024.emnlp-main.202/"
}

@inproceedings{
hu2025signatures,
title={Signatures of human-like processing in Transformer forward passes},
author={Jennifer Hu and Michael A. Lepori and Michael Franke},
booktitle={First Workshop on CogInterp: Interpreting Cognition in Deep Learning Models},
year={2026},
url={https://openreview.net/forum?id=iErDWmZD7y}
}

@article{kuribayashi2025largelanguagemodelshumanlike,
    author = {Kuribayashi, Tatsuki and Oseki, Yohei and Taieb, Souhaib Ben and Inui, Kentaro and Baldwin, Timothy},
    title = {Large Language Models Are Human-Like Internally},
    journal = {TACL},
    volume = {13},
    pages = {1743-1766},
    year = {2025},
    month = {12},
    issn = {2307-387X},
    doi = {10.1162/TACL.a.58},
    url = {https://doi.org/10.1162/TACL.a.58},
    eprint = {https://direct.mit.edu/tacl/article-pdf/doi/10.1162/TACL.a.58/2568514/tacl.a.58.pdf},
}

@INPROCEEDINGS{Arehalli2022-nb,
  title     = "Syntactic surprisal from neural models predicts, but
               underestimates, human processing difficulty from syntactic
               ambiguities",
  author    = "Arehalli, Suhas and Dillon, Brian and Linzen, Tal",
  booktitle = "Proceedings of CoNLL 2022",
  pages     = "301--313",
  month     =  dec,
  year      =  2022,
  url = "https://aclanthology.org/2022.conll-1.20/"
}

@ARTICLE{Timkey2025-cn,
  title    = "Eye movements reveal a dissociation between prediction and
              structural processing in language comprehension",
  author   = "Timkey, William and Huang, Kuan-Jung and Oh, Byung-Doh and Prasad,
              Grusha and Arehalli, Suhas and Linzen, Tal and Dillon, Brian",
  journal  = "PsyArXiv",
  month    =  nov,
  year     =  2025,
  language = "en",
  url = "https://osf.io/preprints/psyarxiv/eq2ra_v1"
  
}

@ARTICLE{Boeve2025-yi,
  title     = "A systematic evaluation of Dutch large language models' surprisal
               estimates in sentence, paragraph and book reading",
  author    = "Boeve, Sam and Bogaerts, Louisa",
  journal   = "Behav. Res. Methods",
  publisher = "Springer Science and Business Media LLC",
  volume    =  57,
  number    =  9,
  pages     =  266,
  month     =  aug,
  year      =  2025,
  language  = "en",
  url     =     "https://pubmed.ncbi.nlm.nih.gov/40825921/"
}

@ARTICLE{Wilks1938-wt,
  title     = "The large-sample distribution of the likelihood ratio for testing
               composite hypotheses",
  author    = "Wilks, S S",
  journal   = "Ann. Math. Stat.",
  publisher = "Institute of Mathematical Statistics",
  volume    =  9,
  number    =  1,
  pages     = "60--62",
  month     =  mar,
  year      =  1938,
  language  = "en",
  url = "https://projecteuclid.org/journals/annals-of-mathematical-statistics/volume-9/issue-1/The-Large-Sample-Distribution-of-the-Likelihood-Ratio-for-Testing/10.1214/aoms/1177732360.full"
}

@book{Cover1999-up,
  title     = "Elements of information theory",
  author    = "Cover, T M",
  publisher = "John Wiley \& Sons",
  year      =  1999,
  url = "https://onlinelibrary.wiley.com/doi/book/10.1002/047174882X"
}

@inproceedings{he-etal-2024-decoding,
    title = "Decoding Probing: Revealing Internal Linguistic Structures in Neural Language Models Using Minimal Pairs",
    author = "He, Linyang  and
      Chen, Peili  and
      Nie, Ercong  and
      Li, Yuanning  and
      Brennan, Jonathan R.",
    editor = "Calzolari, Nicoletta  and
      Kan, Min-Yen  and
      Hoste, Veronique  and
      Lenci, Alessandro  and
      Sakti, Sakriani  and
      Xue, Nianwen",
    booktitle = "Proceedings of LREC-COLING 2024",
    month = may,
    year = "2024",
    url = "https://aclanthology.org/2024.lrec-main.402/",
    pages = "4488--4497"
}

@ARTICLE{Warner1987-wv,
  title     = "Context and distance-to-disambiguation effects in ambiguity
               resolution: Evidence from grammaticality judgments of garden path
               sentences",
  author    = "Warner, John and Glass, Arnold L",
  journal   = "J. Mem. Lang.",
  publisher = "Elsevier BV",
  volume    =  26,
  number    =  6,
  pages     = "714--738",
  month     =  dec,
  year      =  1987,
  language  = "en",
  url = "https://www.sciencedirect.com/science/article/abs/pii/0749596X87901112"
}

@ARTICLE{Levy2008-pl,
  title   = "Modeling the effects of memory on human online sentence processing
             with particle filters",
  author  = "Levy, Roger and Reali, Florencia and Griffiths, Thomas",
  journal = "Proceedings of NIPS 2008",
  volume  =  21,
  year    =  2008,
url = "https://papers.nips.cc/paper_files/paper/2008/hash/a02ffd91ece5e7efeb46db8f10a74059-Abstract.html"
}

@ARTICLE{Itti2009-rk,
  title     = "Bayesian surprise attracts human attention",
  author    = "Itti, Laurent and Baldi, Pierre",
  journal   = "Vision Res.",
  publisher = "Elsevier BV",
  volume    =  49,
  number    =  10,
  pages     = "1295--1306",
  month     =  jun,
  year      =  2009,
  language  = "en",
  url = "https://www.sciencedirect.com/science/article/pii/S0042698908004380"
}

@ARTICLE{Botvinick2001-lb,
  title     = "Conflict monitoring and cognitive control",
  author    = "Botvinick, Matthew M and Braver, Todd S and Barch, Deanna M and
               Carter, Cameron S and Cohen, Jonathan D",
  journal   = "Psychol. Rev.",
  publisher = "American Psychological Association (APA)",
  volume    =  108,
  number    =  3,
  pages     = "624--652",
  month     =  jul,
  year      =  2001,
  language  = "en",
  url = "https://pubmed.ncbi.nlm.nih.gov/11488380/"
}

@ARTICLE{Ness2025-pu,
  title     = "The state of cognitive control in language processing",
  author    = "Ness, Tal and Langlois, Valerie J and Kim, Albert E and Novick,
               Jared M",
  journal   = "Perspect. Psychol. Sci.",
  publisher = "SAGE Publications",
  volume    =  20,
  number    =  2,
  pages     = "219--240",
  month     =  mar,
  year      =  2025,
  language  = "en",
  url    =    "https://pubmed.ncbi.nlm.nih.gov/37819251/"
}

@ARTICLE{Norris2009-pt,
  title     = "Putting it all together: a unified account of word recognition
               and reaction-time distributions",
  author    = "Norris, Dennis",
  journal   = "Psychol. Rev.",
  publisher = "American Psychological Association (APA)",
  volume    =  116,
  number    =  1,
  pages     = "207--219",
  month     =  jan,
  year      =  2009,
  language  = "en"
}

@article{norris2006bayesian,
  title={The Bayesian reader: explaining word recognition as an optimal Bayesian decision process.},
  author={Norris, Dennis},
  journal={Psychological review},
  volume={113},
  number={2},
  pages={327},
  year={2006},
  publisher={American Psychological Association}
}

@ARTICLE{Ratcliff2008-dn,
  title     = "The diffusion decision model: theory and data for two-choice
               decision tasks",
  author    = "Ratcliff, Roger and McKoon, Gail",
  journal   = "Neural Comput.",
  publisher = "MIT Press - Journals",
  volume    =  20,
  number    =  4,
  pages     = "873--922",
  month     =  apr,
  year      =  2008,
  language  = "en"
}

@ARTICLE{Ratcliff1978-dr,
  title     = "A theory of memory retrieval",
  author    = "Ratcliff, Roger",
  journal   = "Psychol. Rev.",
  publisher = "American Psychological Association (APA)",
  volume    =  85,
  number    =  2,
  pages     = "59--108",
  month     =  mar,
  year      =  1978,
  language  = "en"
}

@INPROCEEDINGS{Levy2008-xa,
  title     = "A Noisy-Channel Model of Human Sentence Comprehension under
               Uncertain Input",
  author    = "Levy, Roger",
  booktitle = "Proceedings of EMNLP 2008",
  pages     = "234--243",
  month     =  oct,
  year      =  2008,
  url = "https://aclanthology.org/D08-1025/"
}

@inproceedings{oh-schuler-2025-impact,
    title = "The Impact of Token Granularity on the Predictive Power of Language Model Surprisal",
    author = "Oh, Byung-Doh  and
      Schuler, William",
    editor = "Che, Wanxiang  and
      Nabende, Joyce  and
      Shutova, Ekaterina  and
      Pilehvar, Mohammad Taher",
    booktitle = "Proceedings of the 63rd Annual Meeting of the Association for Computational Linguistics (Volume 1: Long Papers)",
    month = jul,
    year = "2025",
    address = "Vienna, Austria",
    publisher = "Association for Computational Linguistics",
    url = "https://aclanthology.org/2025.acl-long.209/",
    doi = "10.18653/v1/2025.acl-long.209",
    pages = "4150--4162",
    ISBN = "979-8-89176-251-0"
}

@ARTICLE{Barton1993-rz,
  title     = "A case study of anomaly detection: shallow semantic processing
               and cohesion establishment",
  author    = "Barton, S B and Sanford, A J",
  journal   = "Mem. Cognit.",
  publisher = "Springer Science and Business Media LLC",
  volume    =  21,
  number    =  4,
  pages     = "477--487",
  month     =  jul,
  year      =  1993,
  language  = "en"
}

@ARTICLE{Christianson2001-sk,
  title     = "Thematic roles assigned along the garden path linger",
  author    = "Christianson, K and Hollingworth, A and Halliwell, J F and
               Ferreira, F",
  journal   = "Cogn. Psychol.",
  publisher = "Elsevier BV",
  volume    =  42,
  number    =  4,
  pages     = "368--407",
  month     =  jun,
  year      =  2001,
  language  = "en"
}

\clearpage
\appendix

\section*{Appendix}
\label{sec:appendix}

\section{Artifacts}

\subsection{Languauge models}
Table~\ref{tbl:lms} lists the LMs we used.

\begin{table*}[t]
    \centering
    \small
    \begin{tabular}{lp{10cm}l}
    \toprule
       Model & URL & \#params \\
       \cmidrule(lr){1-1} \cmidrule(lr){2-2} \cmidrule(lr){3-3}
       GPT2-small & \url{https://huggingface.co/gpt2} & 117M \\
       GPT2‑medium & \url{https://huggingface.co/gpt2-medium} & 345M \\
       GPT2‑large & \url{https://huggingface.co/gpt2-large} & 774M \\
       GPT2‑xl & \url{https://huggingface.co/gpt2-xl} & 1B \\
       \cmidrule(lr){1-3}
       OPT‑125m & \url{https://huggingface.co/facebook/opt-125m} & 125M \\
       OPT‑1.3b & \url{https://huggingface.co/facebook/opt-1.3b} & 1.3B \\
       OPT‑2.7b & \url{https://huggingface.co/facebook/opt-2.7b} & 2.7B \\
       OPT‑6.7b & \url{https://huggingface.co/facebook/opt-6.7b} & 6.7B \\
       OPT‑13b & \url{https://huggingface.co/facebook/opt-13b} & 13B \\
       OPT‑30b & \url{https://huggingface.co/facebook/opt-30b} & 30B \\
       OPT‑66b & \url{https://huggingface.co/facebook/opt-66b} & 66B \\
       \cmidrule(lr){1-3}
       Pythia‑70m‑deduped & \url{https://huggingface.co/EleutherAI/pythia-70m-deduped} & 70M \\
       Pythia‑160m‑deduped & \url{https://huggingface.co/EleutherAI/pythia-160m-deduped} & 160M \\
       Pythia‑410m‑deduped & \url{https://huggingface.co/EleutherAI/pythia-410m-deduped} & 410M \\
       Pythia‑1b‑deduped & \url{https://huggingface.co/EleutherAI/pythia-1b-deduped} & 1B \\
       Pythia‑1.4b‑deduped & \url{https://huggingface.co/EleutherAI/pythia-1.4b-deduped} & 1.4B \\
       Pythia‑2.8b‑deduped & \url{https://huggingface.co/EleutherAI/pythia-2.8b-deduped} & 2.8B \\
       Pythia‑6.9b‑deduped & \url{https://huggingface.co/EleutherAI/pythia-6.9b-deduped} & 6.9B \\
       Pythia‑12b‑deduped & \url{https://huggingface.co/EleutherAI/pythia-12b-deduped} & 12B \\
        \bottomrule
    \end{tabular}
    \caption{LM details}
    \label{tbl:lms}
\end{table*}

\subsection{Data and tools}
Table~\ref{tab:artifacts} lists data and tools we used.
The experiments used a single NVIDIA RTX 6000 Ada GPU for several hours for surprisal computation.

\begin{table*}[ht!]
    \centering
    \begin{tabular}{p{5cm}p{4.5cm}p{4.5cm}}
    \toprule
        Artifact & License & Usage \\
        \cmidrule(lr){1-1} \cmidrule(lr){2-2} \cmidrule(lr){3-3} 
        Statsmodels~\cite{seabold2010statsmodels} & BSD 3-Clause ``New'' or ``Revised'' License & To train and run regression models \\
        Syntactic Ambiguity Processing Benchmark~\cite{Huang2024-qe} (\url{https://github.com/caplabnyu/sapbenchmark}) & MIT License & To use human reading time data \\
        TunedLens package~\cite{belrose2023eliciting} (\url{https://github.com/AlignmentResearch/tuned-lens}) & MIT License & To compute next-word probabilities \\
        WordFreq~\cite{robyn_speer_2022_7199437} (\url{https://github.com/rspeer/wordfreq}) & Apache 2.0 & To compute unigram frequency of words \\
        Transformers~\cite{wolf2019transformers} (\url{https://github.com/huggingface/transformers}) & Apache 2.0 & To download and run models \\
    \bottomrule
    \end{tabular}
    \caption{Artifacts used in this paper.}
    \label{tab:artifacts}
\end{table*}

\subsection{Reading material statistics}
We posit the assumption that the SAP dataset~\cite{Huang2024-qe} involves more syntactically challenging constructions than other naturalistic reading corpora.
As a quantitative support for this assumption, we computed two general syntactic complexity measures for different corpora: (i) average syntactic tree depth and (ii) average dependency length of a sentence. 
Note that these two scores are normalized by average sentence length, and length control is usually done for a fair inter-corpus comparison~\cite{Futrell2020DependencyOrder}.
These measures are computed with the Spacy \texttt{en\_core\_web\_sm} parser~\cite{spacy}.
In addition to SAP data, we analyzed two other naturalistic reading time corpora of the Natural Stories Corpus (NSC)~\cite{futrell-etal-2018-natural} and the Dundee Corpus~\cite{kennedy2003dundee}.

\begin{table}[t]
    \centering
    \begin{tabular}{lrr}
    \toprule
     Corpus & Tree depth & Dep. length \\ 
     \midrule
     SAP &  \textbf{0.396} & \textbf{0.152} \\
     NSC & 0.302 & 0.139 \\
     Dundee corpus & 0.313 & 0.139 \\
     \bottomrule
    \end{tabular}
    \caption{Corpus statistics on syntactic complexity}
    \label{tab:syn_comp}
\end{table}

Table~\ref{tab:syn_comp} shows the statistics.
These statistics show that SAP is, on average, syntactically more complex.
Note that even in the NSC, one of the relatively syntactically challenging corpus used in naturalistic reading time modeling experiments, including~\citet{kuribayashi2025largelanguagemodelshumanlike}, only 0.02\% sentences, for example, have MVRR ambiguity (see Appendix in~\citet{futrell-etal-2018-natural}); we believe that large-scale targeted datasets such as SAP would still be a reasonable option for our research purpose to obtain statistically reliable results.

It will also be interesting to analyze the effective layer depth even within the naturalistic reading time corpus by dividing it into relatively syntactically challenging data points and others, which will be future work, and our study opens such new analysis directions.

\subsection{Regression models}
\label{app:regression}
In~\cref{sec:exp1} and~\cref{sec:exp2}, we use a linear regression model: 

{\small
\begin{align}
\nonumber
    &\texttt{RT}(w_t) = \beta_0 + \beta_1 \cdot \texttt{Surprisal}(w_t) + \beta_2 \cdot \texttt{Length}(w_t) \\
    \nonumber
    &+ \beta_3 \cdot \texttt{LogFreq}(w_t) + \beta_4 \cdot \texttt{Surprisal}(w_{t-1}) \\
    \nonumber
    &+ \beta_5 \cdot \texttt{Length}(w_{t-1}) + \beta_6 \cdot \texttt{LogFreq}(w_{t-1})  \\
    \nonumber
    &+ \beta_7 \cdot \texttt{Surprisal}(w_{t-2}) + \beta_8 \cdot \texttt{Length}(w_{t-2}) \\
    &+ \beta_9 \cdot \texttt{LogFreq}(w_{t-2}) + \epsilon.
\end{align}
}
For the baseline, we used the regression model without  $\texttt{Surprisal}(w_{t})$ $\texttt{Surprisal}(w_{t-1})$, and $\texttt{Surprisal}(w_{t-2})$.
Word frequency is computed with \texttt{wordfreq} package~\cite{robyn_speer_2022_7199437}, and word length is character-based. Surprisal is computed with an intra-sentential context.

In~\cref{subsec:exp3:analysis}, we analyzed the additive effect of JS to surprisal, where the full model is:

{\small
\begin{align}
\nonumber
    &\texttt{RT}(w_t) = \beta_0 + \beta_1 \cdot \texttt{Surprisal}(w_t) + \beta_2 \cdot \texttt{JS}(w_t) \\
    \nonumber
    &+ \beta_3 \cdot \texttt{Length}(w_t) + \beta_4 \cdot \texttt{LogFreq}(w_t) \\
    \nonumber
    &+ \beta_5 \cdot \texttt{Surprisal}(w_{t-1}) + \beta_6 \cdot \texttt{JS}(w_{t-1}) \\
    \nonumber
    &+ \beta_7 \cdot \texttt{Length}(w_{t-1}) + \beta_8 \cdot \texttt{LogFreq}(w_{t-1}) \\
    \nonumber
    &+ \beta_9 \cdot \texttt{Surprisal}(w_{t-2}) + \beta_{10} \cdot \texttt{JS}(w_{t-2}) \\
    &+ \beta_{11} \cdot \texttt{Length}(w_{t-2}) + \beta_{12} \cdot \texttt{LogFreq}(w_{t-2}) + \epsilon. 
\end{align}
}

\section{Supplimentary results}
\label{appendix:exp}

\subsection{Tuned-lens}
\label{app:tuned-lens}
\begin{figure*}[t]
    \centering
    \begin{subfigure}[b]{\textwidth}
        \centering
        \includegraphics[width=\textwidth]{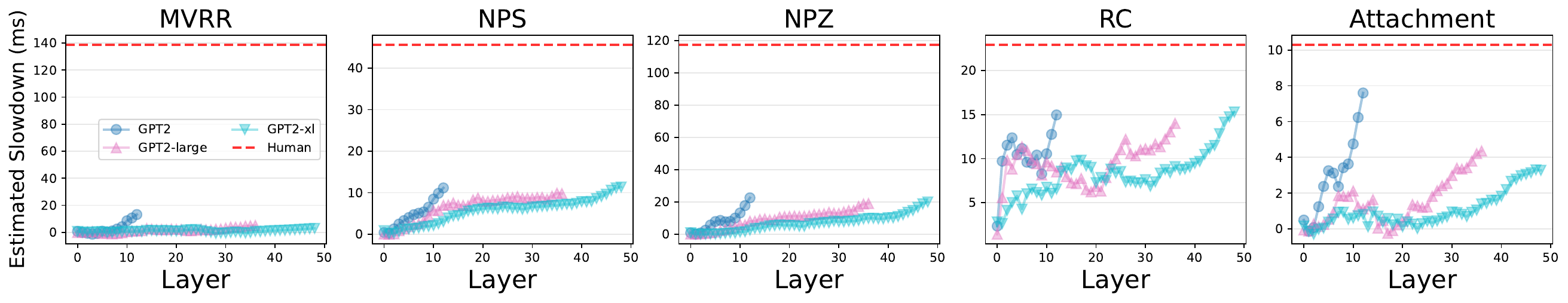}
        \caption{GPT-2 family}
        \label{fig:exp1_gpt2_tuned}
    \end{subfigure}
    
    \begin{subfigure}[b]{\textwidth}
        \centering
        \includegraphics[width=\textwidth]{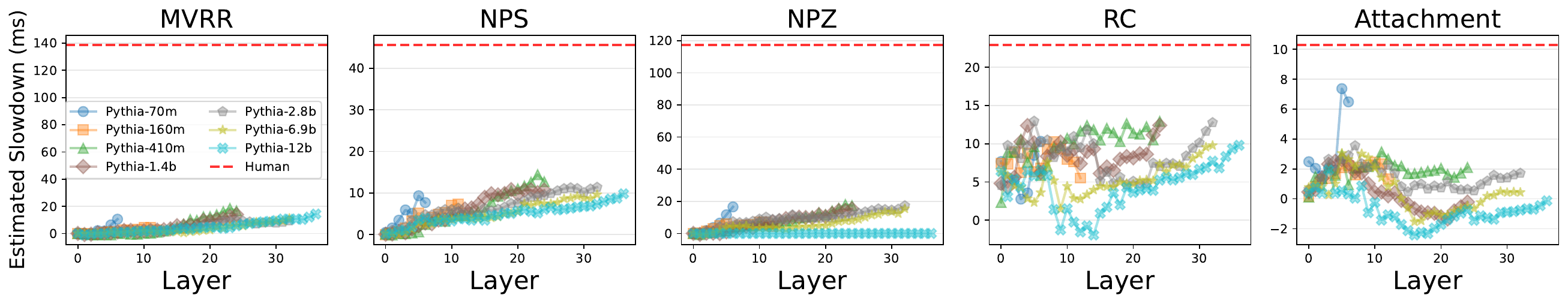}
        \caption{Pythia family}
        \label{fig:exp1_pythia_tuned}
    \end{subfigure}
    
    \begin{subfigure}[b]{\textwidth}
        \centering
        \includegraphics[width=\textwidth]{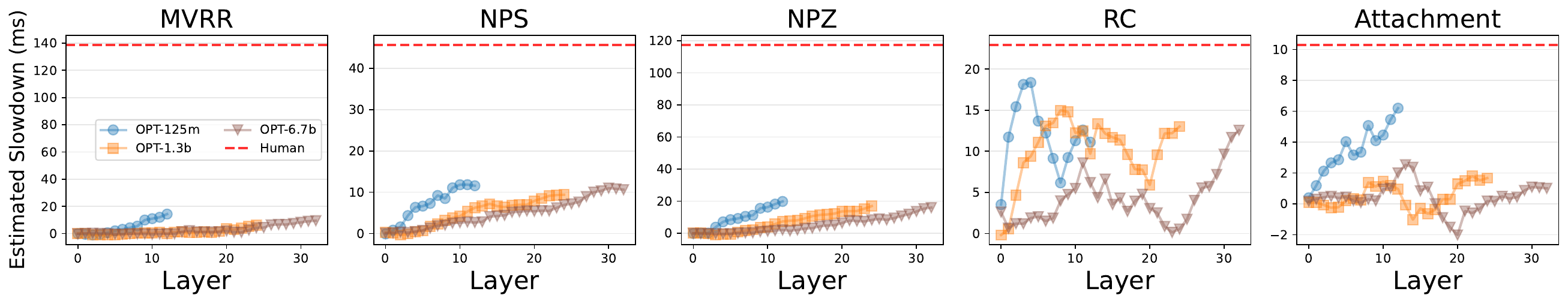}
        \caption{OPT family}
        \label{fig:exp1_opt_tuned}
    \end{subfigure}
    
    \caption{Estimated reading time slowdown by layers for each syntactic construction with TunedLens~\cite{belrose2023eliciting}}
    \label{fig:exp1_2_tuned}
\end{figure*}

We also preliminarily performed the main experiment~\cref{sec:exp1} with TunedLens~\cite{belrose2023eliciting}.
Figure~\ref{fig:exp1_2_tuned} shows the results for models whose tuned-lens parameters are publicly available. 
The patterns generally hold the same as those with LogitLens (Figure~\ref{fig:exp1_2}); no layer can simulate the degree of human reading slowdown.
These results motivate us to focus on experiments with the simpler LogitLens in the subsequent analyses.

\subsection{PPP by probability-update measures}
\label{app:ppp_prob_update}
Tables~\ref{tab:ppp_suprisal_update}, \ref{tab:ppp_kl}, and~\ref{tab:ppp_js} show the model-wise breakdown of PPPs, which are aggregated in Figure~\ref{fig:fig_kl}.

\subsection{Slowdown estimates by probability-update measures}
\label{app:effect_prob_update}
Tables~\ref{tab:effect_surprisal_update}, \ref{tab:effect_kl}, and~\ref{tab:effect_js} show the estimated reading time slowdowns for t* and t*+1 data points, following the method of~\cref{sec:exp1}.

\subsection{Additive effect of JS to surprisal}
\label{app:ppp_prob_update_additive}
Table~\ref{tab:surprisal_plus_js} shows the model-wise breakdown of the additive effect of the JS value on top of surprisal.
For the log likelihood ratio test, we compare the likelihood of two nested models: (i) the above full model vs. (ii) the model without JS factors (but still with surprisals). 
That is, the difference in degrees of freedom between the two models is three.

\begin{table}[t]
\centering
\footnotesize
\tabcolsep=0.002cm
\begin{tabular}{lrrrrrrrrrrrrrrr}
\toprule
& \multicolumn{2}{c}{MVRR} & \multicolumn{2}{c}{NPS} & \multicolumn{2}{c}{NPZ} & \multicolumn{2}{c}{RC} & \multicolumn{2}{c}{Att.} \\
\cmidrule(r){2-3} \cmidrule(r){4-5} \cmidrule(r){6-7} \cmidrule(r){8-9} \cmidrule(r){10-11}
Model  & \multicolumn{1}{c}{All} & \multicolumn{1}{c}{\textsc{RoI}} & \multicolumn{1}{c}{All} & \multicolumn{1}{c}{\textsc{RoI}} & \multicolumn{1}{c}{All} & \multicolumn{1}{c}{\textsc{RoI}} & \multicolumn{1}{c}{All} & \multicolumn{1}{c}{\textsc{RoI}} & \multicolumn{1}{c}{All} & \multicolumn{1}{c}{\textsc{RoI}} \\
\midrule
G2-sm  & 11.7$^\dagger$ & 14.1$^\dagger$ & 1.1$^{\phantom{\dagger}}$ & 13.8$^\dagger$ & 7.8$^\dagger$ & 24.5$^\dagger$ & 12.0$^\dagger$ & 1.7$^{\phantom{\dagger}}$ & 4.5$^\dagger$ & 0.8$^{\phantom{\dagger}}$ \\
G2-md & 10.8$^\dagger$ & 8.7$^\dagger$  & 0.8$^{\phantom{\dagger}}$ & 6.9$^\dagger$  & 12.5$^\dagger$ & 26.3$^\dagger$ & 13.6$^\dagger$ & 1.5$^{\phantom{\dagger}}$ & 4.9$^\dagger$ & 1.9$^{\phantom{\dagger}}$ \\
G2-lg  & 11.3$^\dagger$ & 7.0$^\dagger$  & 0.7$^{\phantom{\dagger}}$ & 10.3$^\dagger$ & 10.5$^\dagger$ & 27.0$^\dagger$ & 14.1$^\dagger$ & 1.8$^{\phantom{\dagger}}$ & 3.2$^{\phantom{\dagger}}$ & 2.2$^{\phantom{\dagger}}$ \\
G2-xl  & 11.4$^\dagger$ & 5.5$^\dagger$  & 0.8$^{\phantom{\dagger}}$ & 7.3$^\dagger$  & 10.5$^\dagger$ & 28.1$^\dagger$ & 12.3$^\dagger$ & 1.6$^{\phantom{\dagger}}$ & 4.2$^\dagger$ & 2.1$^{\phantom{\dagger}}$ \\
\cmidrule(r){1-1} \cmidrule(r){2-3} \cmidrule(r){4-5} \cmidrule(r){6-7} \cmidrule(r){8-9} \cmidrule(r){10-11}
OT-125m & 11.5$^\dagger$ & 13.5$^\dagger$ & 0.4$^{\phantom{\dagger}}$ & 11.9$^\dagger$ & 3.8$^{\phantom{\dagger}}$ & 20.1$^\dagger$ & 11.6$^\dagger$ & 1.9$^{\phantom{\dagger}}$ & 4.5$^\dagger$ & 1.1$^{\phantom{\dagger}}$ \\
OT-1.3b & 7.8$^\dagger$  & 7.8$^\dagger$  & 0.6$^{\phantom{\dagger}}$ & 8.2$^\dagger$  & 6.8$^\dagger$ & 20.9$^\dagger$ & 12.7$^\dagger$ & 2.1$^{\phantom{\dagger}}$ & 4.8$^\dagger$ & 1.9$^{\phantom{\dagger}}$ \\
OT-2.7b & 8.3$^\dagger$  & 11.0$^\dagger$ & 0.5$^{\phantom{\dagger}}$ & 7.0$^\dagger$  & 6.3$^\dagger$ & 20.0$^\dagger$ & 11.1$^\dagger$ & 1.7$^{\phantom{\dagger}}$ & 4.9$^\dagger$ & 1.7$^{\phantom{\dagger}}$ \\
OT-6.7b & 8.2$^\dagger$  & 7.1$^\dagger$  & 0.8$^{\phantom{\dagger}}$ & 8.3$^\dagger$  & 5.9$^\dagger$ & 19.0$^\dagger$ & 10.1$^\dagger$ & 1.3$^{\phantom{\dagger}}$ & 5.6$^\dagger$ & 1.8$^{\phantom{\dagger}}$ \\
OT-13b  & 8.4$^\dagger$  & 11.1$^\dagger$ & 0.4$^{\phantom{\dagger}}$ & 6.6$^\dagger$  & 5.5$^\dagger$ & 17.8$^\dagger$ & 8.4$^\dagger$  & 2.6$^{\phantom{\dagger}}$ & 7.3$^\dagger$ & 1.8$^{\phantom{\dagger}}$ \\
OT-30b  & 6.7$^\dagger$  & 3.1$^{\phantom{\dagger}}$ & 0.7$^{\phantom{\dagger}}$ & 3.7$^{\phantom{\dagger}}$ & 5.9$^\dagger$ & 10.5$^\dagger$ & 7.3$^\dagger$  & 2.0$^{\phantom{\dagger}}$ & 7.8$^\dagger$ & 1.4$^{\phantom{\dagger}}$ \\
OT-66b  & 7.9$^\dagger$  & 7.7$^\dagger$  & 0.6$^{\phantom{\dagger}}$ & 5.1$^\dagger$  & 4.3$^\dagger$ & 15.3$^\dagger$ & 10.7$^\dagger$ & 1.8$^{\phantom{\dagger}}$ & 6.9$^\dagger$ & 1.1$^{\phantom{\dagger}}$ \\
\cmidrule(r){1-1} \cmidrule(r){2-3} \cmidrule(r){4-5} \cmidrule(r){6-7} \cmidrule(r){8-9} \cmidrule(r){10-11}
PT-70m  & 6.8$^\dagger$ & 2.6$^{\phantom{\dagger}}$ & 1.6$^{\phantom{\dagger}}$ & 1.5$^{\phantom{\dagger}}$ & 3.5$^{\phantom{\dagger}}$ & 8.7$^\dagger$ & 9.6$^\dagger$ & 0.3$^{\phantom{\dagger}}$ & 2.4$^{\phantom{\dagger}}$ & 0.6$^{\phantom{\dagger}}$ \\
PT-160m & 7.8$^\dagger$ & 1.6$^{\phantom{\dagger}}$ & 1.5$^{\phantom{\dagger}}$ & 1.6$^{\phantom{\dagger}}$ & 11.3$^\dagger$ & 0.9$^{\phantom{\dagger}}$ & 6.7$^\dagger$ & 0.7$^{\phantom{\dagger}}$ & 5.8$^\dagger$ & 4.2$^\dagger$ \\
PT-410m & 8.9$^\dagger$ & 8.8$^\dagger$ & 0.1$^{\phantom{\dagger}}$ & 11.9$^\dagger$ & 2.4$^{\phantom{\dagger}}$ & 14.8$^\dagger$ & 14.0$^\dagger$ & 0.8$^{\phantom{\dagger}}$ & 9.3$^\dagger$ & 0.9$^{\phantom{\dagger}}$ \\
PT-1b   & 9.5$^\dagger$ & 8.7$^\dagger$ & 0.6$^{\phantom{\dagger}}$ & 8.5$^\dagger$ & 1.8$^{\phantom{\dagger}}$ & 6.2$^\dagger$ & 9.1$^\dagger$ & 0.7$^{\phantom{\dagger}}$ & 2.9$^{\phantom{\dagger}}$ & 1.7$^{\phantom{\dagger}}$ \\
PT-1.4b & 6.0$^\dagger$ & 9.1$^\dagger$ & 1.5$^{\phantom{\dagger}}$ & 6.6$^\dagger$ & 1.1$^{\phantom{\dagger}}$ & 11.1$^\dagger$ & 7.0$^\dagger$ & 0.3$^{\phantom{\dagger}}$ & 6.2$^\dagger$ & 2.2$^{\phantom{\dagger}}$ \\
PT-2.8b & 5.8$^\dagger$ & 3.7$^{\phantom{\dagger}}$ & 0.1$^{\phantom{\dagger}}$ & 7.3$^\dagger$ & 4.3$^\dagger$ & 13.8$^\dagger$ & 6.9$^\dagger$ & 0.3$^{\phantom{\dagger}}$ & 4.8$^\dagger$ & 1.2$^{\phantom{\dagger}}$ \\
PT-6.9b & 8.5$^\dagger$ & 8.6$^\dagger$ & 2.2$^{\phantom{\dagger}}$ & 10.8$^\dagger$ & 1.9$^{\phantom{\dagger}}$ & 15.0$^\dagger$ & 11.6$^\dagger$ & 0.6$^{\phantom{\dagger}}$ & 8.9$^\dagger$ & 1.5$^{\phantom{\dagger}}$ \\
PT-12b  & 6.8$^\dagger$ & 10.4$^\dagger$ & 1.0$^{\phantom{\dagger}}$ & 9.8$^\dagger$ & 2.7$^{\phantom{\dagger}}$ & 10.8$^\dagger$ & 8.2$^\dagger$ & 0.3$^{\phantom{\dagger}}$ & 10.8$^\dagger$ & 1.8$^{\phantom{\dagger}}$ \\
\bottomrule
\end{tabular}
\caption{PPP ($\Delta$LL) of surprisal update by model, phenomenon, and data group. The value with $^\dagger$ indicates statistical significance ($p < 0.05$) with a likelihood ratio test.}
\label{tab:ppp_suprisal_update}
\end{table}

\begin{table}[t]
\centering
\footnotesize
\tabcolsep=0.002cm
\begin{tabular}{lllllllllll}
\toprule
 & \multicolumn{2}{c}{MVRR} & \multicolumn{2}{c}{NPS} & \multicolumn{2}{c}{NPZ} & \multicolumn{2}{c}{RC} & \multicolumn{2}{c}{Att.} \\
 \cmidrule(r){2-3} \cmidrule(r){4-5} \cmidrule(r){6-7} \cmidrule(r){8-9} \cmidrule(r){10-11}
Model  & \multicolumn{1}{c}{All} & \multicolumn{1}{c}{\textsc{RoI}} & \multicolumn{1}{c}{All} & \multicolumn{1}{c}{\textsc{RoI}} & \multicolumn{1}{c}{All} & \multicolumn{1}{c}{\textsc{RoI}} & \multicolumn{1}{c}{All} & \multicolumn{1}{c}{\textsc{RoI}} & \multicolumn{1}{c}{All} & \multicolumn{1}{c}{\textsc{RoI}} \\
\midrule
G2-sm & 10.0$^\dagger$ & 2.5$^{\phantom{\dagger}}$ & 0.4$^{\phantom{\dagger}}$ & 0.7$^{\phantom{\dagger}}$ & 1.7$^{\phantom{\dagger}}$ & 9.7$^\dagger$ & 11.2$^\dagger$ & 0.6$^{\phantom{\dagger}}$ & 5.3$^\dagger$ & 2.6$^{\phantom{\dagger}}$ \\
G2-md & 9.7$^\dagger$ & 2.0$^{\phantom{\dagger}}$ & 0.8$^{\phantom{\dagger}}$ & 5.1$^\dagger$ & 2.6$^{\phantom{\dagger}}$ & 15.0$^\dagger$ & 18.9$^\dagger$ & 0.2$^{\phantom{\dagger}}$ & 20.9$^\dagger$ & 4.9$^\dagger$ \\
G2-lg & 7.4$^\dagger$ & 2.2$^{\phantom{\dagger}}$ & 0.5$^{\phantom{\dagger}}$ & 0.8$^{\phantom{\dagger}}$ & 1.8$^{\phantom{\dagger}}$ & 16.0$^\dagger$ & 16.5$^\dagger$ & 0.1$^{\phantom{\dagger}}$ & 16.3$^\dagger$ & 7.4$^\dagger$ \\
G2-xl & 3.1$^{\phantom{\dagger}}$ & 2.2$^{\phantom{\dagger}}$ & 1.0$^{\phantom{\dagger}}$ & 0.3$^{\phantom{\dagger}}$ & 1.4$^{\phantom{\dagger}}$ & 15.5$^\dagger$ & 15.5$^\dagger$ & 0.4$^{\phantom{\dagger}}$ & 16.2$^\dagger$ & 4.3$^\dagger$ \\
\cmidrule(r){1-1} \cmidrule(r){2-3} \cmidrule(r){4-5} \cmidrule(r){6-7} \cmidrule(r){8-9} \cmidrule(r){10-11}
OT-125m & 7.7$^\dagger$ & 0.2$^{\phantom{\dagger}}$ & 0.4$^{\phantom{\dagger}}$ & 2.1$^{\phantom{\dagger}}$ & 2.7$^{\phantom{\dagger}}$ & 8.5$^\dagger$ & 8.5$^\dagger$ & 2.4$^{\phantom{\dagger}}$ & 19.1$^\dagger$ & 6.0$^\dagger$ \\
OT-1.3b & 9.6$^\dagger$ & 1.6$^{\phantom{\dagger}}$ & 2.6$^{\phantom{\dagger}}$ & 1.4$^{\phantom{\dagger}}$ & 4.3$^\dagger$ & 13.3$^\dagger$ & 15.4$^\dagger$ & 0.5$^{\phantom{\dagger}}$ & 34.8$^\dagger$ & 4.0$^\dagger$ \\
OT-2.7b & 9.5$^\dagger$ & 1.9$^{\phantom{\dagger}}$ & 2.1$^{\phantom{\dagger}}$ & 1.2$^{\phantom{\dagger}}$ & 4.4$^\dagger$ & 10.8$^\dagger$ & 11.5$^\dagger$ & 0.6$^{\phantom{\dagger}}$ & 34.3$^\dagger$ & 3.7$^{\phantom{\dagger}}$ \\
OT-6.7b & 2.7$^{\phantom{\dagger}}$ & 2.9$^{\phantom{\dagger}}$ & 0.8$^{\phantom{\dagger}}$ & 4.0$^\dagger$ & 1.1$^{\phantom{\dagger}}$ & 8.9$^\dagger$ & 10.2$^\dagger$ & 0.7$^{\phantom{\dagger}}$ & 24.6$^\dagger$ & 2.4$^{\phantom{\dagger}}$ \\
OT-13b & 6.5$^\dagger$ & 4.6$^\dagger$ & 1.7$^{\phantom{\dagger}}$ & 2.8$^{\phantom{\dagger}}$ & 3.8$^{\phantom{\dagger}}$ & 8.2$^\dagger$ & 17.5$^\dagger$ & 0.7$^{\phantom{\dagger}}$ & 26.4$^\dagger$ & 2.9$^{\phantom{\dagger}}$ \\
OT-30b & 4.8$^\dagger$ & 5.9$^\dagger$ & 1.3$^{\phantom{\dagger}}$ & 7.4$^\dagger$ & 3.1$^{\phantom{\dagger}}$ & 7.1$^\dagger$ & 17.0$^\dagger$ & 0.6$^{\phantom{\dagger}}$ & 31.4$^\dagger$ & 3.3$^{\phantom{\dagger}}$ \\
OT-66b & 2.2$^{\phantom{\dagger}}$ & 1.7$^{\phantom{\dagger}}$ & 0.7$^{\phantom{\dagger}}$ & 1.2$^{\phantom{\dagger}}$ & 2.9$^{\phantom{\dagger}}$ & 12.1$^\dagger$ & 9.3$^\dagger$ & 1.2$^{\phantom{\dagger}}$ & 17.5$^\dagger$ & 3.5$^{\phantom{\dagger}}$ \\
\cmidrule(r){1-1} \cmidrule(r){2-3} \cmidrule(r){4-5} \cmidrule(r){6-7} \cmidrule(r){8-9} \cmidrule(r){10-11}
PT-70m & 10.6$^\dagger$ & 2.0$^{\phantom{\dagger}}$ & 1.6$^{\phantom{\dagger}}$ & 1.4$^{\phantom{\dagger}}$ & 0.7$^{\phantom{\dagger}}$ & 0.8$^{\phantom{\dagger}}$ & 22.5$^\dagger$ & 3.3$^{\phantom{\dagger}}$ & 22.6$^\dagger$ & 7.0$^\dagger$ \\
PT-160m & 24.5$^\dagger$ & 0.5$^{\phantom{\dagger}}$ & 3.1$^{\phantom{\dagger}}$ & 2.6$^{\phantom{\dagger}}$ & 26.1$^\dagger$ & 1.8$^{\phantom{\dagger}}$ & 50.3$^\dagger$ & 3.8$^{\phantom{\dagger}}$ & 35.1$^\dagger$ & 9.7$^\dagger$ \\
PT-410m & 9.9$^\dagger$ & 2.9$^{\phantom{\dagger}}$ & 1.2$^{\phantom{\dagger}}$ & 1.7$^{\phantom{\dagger}}$ & 0.6$^{\phantom{\dagger}}$ & 2.4$^{\phantom{\dagger}}$ & 9.2$^\dagger$ & 1.2$^{\phantom{\dagger}}$ & 26.7$^\dagger$ & 9.2$^\dagger$ \\
PT-1b & 38.5$^\dagger$ & 3.2$^{\phantom{\dagger}}$ & 1.9$^{\phantom{\dagger}}$ & 1.9$^{\phantom{\dagger}}$ & 8.6$^\dagger$ & 19.6$^\dagger$ & 34.8$^\dagger$ & 2.1$^{\phantom{\dagger}}$ & 22.3$^\dagger$ & 11.7$^\dagger$ \\
PT-1.4b & 7.6$^\dagger$ & 0.3$^{\phantom{\dagger}}$ & 0.1$^{\phantom{\dagger}}$ & 1.9$^{\phantom{\dagger}}$ & 1.1$^{\phantom{\dagger}}$ & 5.9$^\dagger$ & 14.9$^\dagger$ & 1.8$^{\phantom{\dagger}}$ & 16.0$^\dagger$ & 6.2$^\dagger$ \\
PT-2.8b & 20.3$^\dagger$ & 0.7$^{\phantom{\dagger}}$ & 1.4$^{\phantom{\dagger}}$ & 3.3$^{\phantom{\dagger}}$ & 6.1$^\dagger$ & 12.7$^\dagger$ & 38.0$^\dagger$ & 3.7$^{\phantom{\dagger}}$ & 31.7$^\dagger$ & 5.0$^\dagger$ \\
PT-6.9b & 4.3$^\dagger$ & 0.2$^{\phantom{\dagger}}$ & 0.9$^{\phantom{\dagger}}$ & 2.2$^{\phantom{\dagger}}$ & 1.8$^{\phantom{\dagger}}$ & 8.8$^\dagger$ & 18.7$^\dagger$ & 2.2$^{\phantom{\dagger}}$ & 21.6$^\dagger$ & 6.6$^\dagger$ \\
PT-12b & 4.5$^\dagger$ & 0.3$^{\phantom{\dagger}}$ & 0.2$^{\phantom{\dagger}}$ & 1.3$^{\phantom{\dagger}}$ & 1.5$^{\phantom{\dagger}}$ & 8.0$^\dagger$ & 18.2$^\dagger$ & 2.3$^{\phantom{\dagger}}$ & 21.9$^\dagger$ & 5.5$^\dagger$ \\
\bottomrule
\end{tabular}
\caption{PPP ($\Delta$LL) of KL-divergence $\mathrm{KL}(Q || P)$ by model, phenomenon, and data group. The value with $^\dagger$ indicates statistical significance ($p < 0.05$) with a likelihood ratio test.}
\label{tab:ppp_kl}
\end{table}

\begin{table}[t]
\centering
\footnotesize
\tabcolsep=0.002cm
\begin{tabular}{lllllllllll}
\toprule
 & \multicolumn{2}{c}{MVRR} & \multicolumn{2}{c}{NPS} & \multicolumn{2}{c}{NPZ} & \multicolumn{2}{c}{RC} & \multicolumn{2}{c}{Att.} \\
 \cmidrule(r){2-3} \cmidrule(r){4-5} \cmidrule(r){6-7} \cmidrule(r){8-9} \cmidrule(r){10-11}
Model  & \multicolumn{1}{c}{All} & \multicolumn{1}{c}{\textsc{RoI}} & \multicolumn{1}{c}{All} & \multicolumn{1}{c}{\textsc{RoI}} & \multicolumn{1}{c}{All} & \multicolumn{1}{c}{\textsc{RoI}} & \multicolumn{1}{c}{All} & \multicolumn{1}{c}{\textsc{RoI}} & \multicolumn{1}{c}{All} & \multicolumn{1}{c}{\textsc{RoI}} \\
\midrule
G2-sm & 10.4$^\dagger$ & 2.4 & 0.4 & 2.5 & 2.2 & 12.3$^\dagger$ & 15.4$^\dagger$ & 0.5 & 17.2$^\dagger$ & 3.5 \\
G2-md & 17.2$^\dagger$ & 3.0 & 2.5 & 2.8 & 6.6$^\dagger$ & 13.9$^\dagger$ & 30.8$^\dagger$ & 0.4 & 32.2$^\dagger$ & 6.5$^\dagger$ \\
G2-lg & 9.5$^\dagger$ & 5.2$^\dagger$ & 1.6 & 1.3 & 1.9 & 12.0$^\dagger$ & 19.9$^\dagger$ & 0.1 & 27.4$^\dagger$ & 5.1$^\dagger$ \\
G2-xl & 7.7$^\dagger$ & 2.6 & 1.5 & 0.9 & 2.2 & 11.9$^\dagger$ & 22.4$^\dagger$ & 0.3 & 29.5$^\dagger$ & 5.0$^\dagger$ \\
\cmidrule(r){1-1} \cmidrule(r){2-3} \cmidrule(r){4-5} \cmidrule(r){6-7} \cmidrule(r){8-9} \cmidrule(r){10-11}
OT-125m & 33.7$^\dagger$ & 0.8 & 2.1 & 2.3 & 2.2 & 8.1$^\dagger$ & 23.6$^\dagger$ & 0.5 & 29.5$^\dagger$ & 5.3$^\dagger$ \\
OT-1.3b & 12.7$^\dagger$ & 2.1 & 2.6 & 0.4 & 6.2$^\dagger$ & 10.9$^\dagger$ & 30.1$^\dagger$ & 0.7 & 30.7$^\dagger$ & 4.6$^\dagger$ \\
OT-2.7b & 12.1$^\dagger$ & 1.6 & 2.0 & 0.9 & 6.7$^\dagger$ & 9.8$^\dagger$ & 30.1$^\dagger$ & 1.0 & 30.3$^\dagger$ & 4.3$^\dagger$ \\
OT-6.7b & 6.1$^\dagger$ & 3.4 & 1.1 & 3.4 & 3.1 & 13.3$^\dagger$ & 24.1$^\dagger$ & 0.9 & 25.8$^\dagger$ & 3.1 \\
OT-13b & 9.1$^\dagger$ & 4.7$^\dagger$ & 1.6 & 2.1 & 5.2$^\dagger$ & 9.2$^\dagger$ & 28.5$^\dagger$ & 0.8 & 28.7$^\dagger$ & 3.4 \\
OT-30b & 7.1$^\dagger$ & 8.8$^\dagger$ & 1.0 & 6.5$^\dagger$ & 4.3$^\dagger$ & 14.2$^\dagger$ & 25.0$^\dagger$ & 0.7 & 28.4$^\dagger$ & 3.3 \\
OT-66b & 2.4 & 2.6 & 0.2 & 1.7 & 2.5 & 9.7$^\dagger$ & 11.6$^\dagger$ & 1.3 & 16.0$^\dagger$ & 3.8 \\
\cmidrule(r){1-1} \cmidrule(r){2-3} \cmidrule(r){4-5} \cmidrule(r){6-7} \cmidrule(r){8-9} \cmidrule(r){10-11}
PT-70m & 24.4$^\dagger$ & 0.6 & 1.8 & 1.5 & 3.7 & 6.0$^\dagger$ & 28.5$^\dagger$ & 3.5 & 33.5$^\dagger$ & 8.8$^\dagger$ \\
PT-160m & 20.9$^\dagger$ & 0.2 & 2.3 & 4.0$^\dagger$ & 3.7 & 6.9$^\dagger$ & 26.0$^\dagger$ & 3.1 & 30.9$^\dagger$ & 7.3$^\dagger$ \\
PT-410m & 20.8$^\dagger$ & 1.5 & 1.7 & 2.2 & 1.6 & 5.3$^\dagger$ & 19.9$^\dagger$ & 2.3 & 28.6$^\dagger$ & 5.3$^\dagger$ \\
PT-1b & 18.9$^\dagger$ & 0.6 & 1.8 & 2.1 & 3.0 & 7.4$^\dagger$ & 26.5$^\dagger$ & 2.4 & 32.4$^\dagger$ & 7.9$^\dagger$ \\
PT-1.4b & 14.4$^\dagger$ & 0.7 & 0.9 & 4.0$^\dagger$ & 1.5 & 5.9$^\dagger$ & 19.8$^\dagger$ & 1.8 & 24.6$^\dagger$ & 7.2$^\dagger$ \\
PT-2.8b & 22.2$^\dagger$ & 0.4 & 2.3 & 2.7 & 3.9 & 5.8$^\dagger$ & 30.4$^\dagger$ & 3.4 & 32.5$^\dagger$ & 7.7$^\dagger$ \\
PT-6.9b & 8.7$^\dagger$ & 0.9 & 1.1 & 0.4 & 1.8 & 5.4$^\dagger$ & 22.4$^\dagger$ & 1.9 & 23.8$^\dagger$ & 7.1$^\dagger$ \\
PT-12b & 12.9$^\dagger$ & 0.4 & 1.0 & 1.1 & 1.7 & 5.1$^\dagger$ & 23.7$^\dagger$ & 2.3 & 25.8$^\dagger$ & 7.5$^\dagger$ \\
\bottomrule
\end{tabular}
\caption{PPP ($\Delta$LL) of JS-divergence by model, phenomenon, and data group. The value with $^\dagger$ indicates statistical significance ($p < 0.05$) with a likelihood ratio test.}
\label{tab:ppp_js}
\end{table}

\begin{table}
\centering
\small
\tabcolsep=0.1cm
\begin{tabular}{lrrrrr}
\toprule
Models & MVRR & NPS & NPZ & RC & Attach. \\
\midrule
GPT2-small & 5.00 & 4.30 & 7.97 & 5.91 & 2.61 \\
GPT2-medium & 2.13 & 2.25 & 5.40 & 5.29 & 1.55 \\
GPT2-large & 1.18 & 1.30 & 3.66 & 5.02 & 1.01 \\
GPT2-xl & $-$0.27 & 1.14 & 2.52 & 5.16 & 0.34 \\
\cmidrule(r){2-2} \cmidrule(r){3-3} \cmidrule(r){4-4} \cmidrule(r){5-5} \cmidrule(r){6-6}
OPT-125m & 3.41 & 2.53 & 2.41 & 4.84 & 1.06 \\
OPT-1.3b & 0.01 & 0.80 & 0.77 & 3.99 & $-$0.43 \\
OPT-2.7b & 0.23 & 0.79 & 0.78 & 2.70 & $-$0.62 \\
OPT-6.7b & 1.39 & 2.29 & 2.72 & 3.86 & $-$0.38 \\
OPT-13b & $-$0.20 & 1.09 & 0.64 & 2.33 & $-$0.80 \\
OPT-30b & $-$0.41 & 1.98 & 1.87 & 2.58 & $-$1.11 \\
OPT-66b & $-$0.83 & $-$0.38 & $-$1.56 & 0.38 & $-$0.70 \\
\cmidrule(r){2-2} \cmidrule(r){3-3} \cmidrule(r){4-4} \cmidrule(r){5-5} \cmidrule(r){6-6}
Pythia-70m & 1.36 & 2.14 & 3.70 & 0.04 & 1.26 \\
Pythia-160m & 0.59 & 0.65 & 3.06 & $-$3.73 & $-$0.10 \\
Pythia-410m & $-$2.18 & 0.28 & $-$0.33 & $-$0.30 & $-$1.74 \\
Pythia-1b & $-$2.80 & $-$1.90 & $-$2.59 & $-$0.79 & $-$0.55 \\
Pythia-1.4b & $-$2.67 & $-$0.92 & $-$2.83 & $-$1.01 & $-$0.59 \\
Pythia-2.8b & $-$1.91 & 0.16 & $-$1.11 & $-$0.58 & $-$1.61 \\
Pythia-6.9b & $-$1.51 & $-$0.03 & $-$1.38 & $-$1.21 & $-$1.07 \\
Pythia-12b & $-$0.80 & 1.02 & $-$0.25 & $-$1.22 & $-$1.03 \\
\bottomrule
\end{tabular}
\caption{Estimated slowdown by surprisal-update}
\label{tab:effect_surprisal_update}
\end{table}

\begin{table}
\centering
\small
\tabcolsep=0.1cm
\begin{tabular}{lrrrrr}
\toprule
Models & MVRR & NPS & NPZ & RC & Attach. \\
\midrule
GPT2 & 0.07 & $-$0.51 & $-$0.86 & 3.36 & 0.08 \\
GPT2-medium & $-$0.34 & $-$1.42 & $-$1.85 & 0.52 & 0.06 \\
GPT2-large & 0.20 & $-$0.63 & 0.46 & 1.85 & 0.29 \\
GPT2-xl & $-$0.31 & $-$0.42 & 0.47 & 0.81 & 0.55 \\
\cmidrule(r){2-2} \cmidrule(r){3-3} \cmidrule(r){4-4} \cmidrule(r){5-5} \cmidrule(r){6-6}
OPT-125m & 0.08 & $-$0.33 & 0.14 & $-$0.86 & $-$0.05 \\
OPT-1.3b & $-$0.39 & $-$1.12 & $-$1.16 & 0.08 & $-$0.23 \\
OPT-2.7b & 0.26 & $-$1.23 & $-$0.83 & 0.27 & $-$0.01 \\
OPT-6.7b & 0.44 & $-$0.85 & $-$1.06 & $-$0.98 & 0.02 \\
OPT-13b & $-$0.60 & $-$1.77 & $-$0.68 & $-$0.86 & $-$0.15 \\
OPT-30b & $-$1.08 & $-$1.76 & $-$1.58 & $-$0.49 & $-$0.15 \\
OPT-66b & 0.85 & $-$0.58 & $-$0.54 & $-$1.27 & $-$0.13 \\
\cmidrule(r){2-2} \cmidrule(r){3-3} \cmidrule(r){4-4} \cmidrule(r){5-5} \cmidrule(r){6-6}
Pythia-70m & 2.36 & 2.20 & 1.52 & $-$0.10 & 0.38 \\
Pythia-160m & 0.50 & 0.91 & 0.32 & $-$3.01 & 0.22 \\
Pythia-410m & 0.98 & $-$0.65 & $-$1.09 & $-$0.77 & $-$0.26 \\
Pythia-1b & 3.83 & 1.03 & 4.15 & $-$0.19 & 0.30 \\
Pythia-1.4b & 2.07 & 0.77 & 0.50 & $-$0.34 & $-$0.31 \\
Pythia-2.8b & 1.98 & 0.69 & 3.84 & $-$2.67 & $-$0.29 \\
Pythia-6.9b & 1.21 & $-$0.19 & $-$0.67 & $-$0.99 & $-$0.66 \\
Pythia-12b & 2.24 & $-$0.39 & $-$0.19 & $-$1.18 & $-$0.83 \\
\bottomrule
\end{tabular}
\caption{Estimated slowdown by KL-divergence $\mathrm{KL}(Q || P)$}
\label{tab:effect_kl}
\end{table}

\begin{table}
\centering
\small
\tabcolsep=0.1cm
\begin{tabular}{lrrrrr}
\toprule
Models & MVRR & NPS & NPZ & RC & Attach. \\
\midrule
GPT2 & $-$0.06 & $-$0.56 & $-$0.59 & 2.97 & 0.19 \\
GPT2-medium & $-$0.59 & 0.28 & 0.38 & 0.69 & 0.15 \\
GPT2-large & $-$0.58 & $-$0.33 & 0.08 & 1.27 & 0.14 \\
GPT2-xl & $-$0.40 & 0.23 & 0.31 & 0.58 & 0.49 \\
\cmidrule(r){2-2} \cmidrule(r){3-3} \cmidrule(r){4-4} \cmidrule(r){5-5} \cmidrule(r){6-6}
OPT-125m & 0.04 & $-$0.01 & $-$0.03 & 0.65 & 0.09 \\
OPT-1.3b & $-$0.39 & $-$0.28 & 0.27 & $-$0.08 & $-$0.09 \\
OPT-2.7b & 0.09 & $-$0.98 & 0.06 & 0.23 & $-$0.04 \\
OPT-6.7b & 0.22 & $-$1.24 & $-$1.26 & $-$0.28 & $-$0.13 \\
OPT-13b & $-$0.53 & $-$1.21 & 0.03 & $-$0.81 & $-$0.09 \\
OPT-30b & $-$1.56 & $-$2.70 & $-$1.81 & $-$0.45 & $-$0.15 \\
OPT-66b & 0.53 & $-$0.80 & $-$0.65 & $-$1.36 & $-$0.10 \\
\cmidrule(r){2-2} \cmidrule(r){3-3} \cmidrule(r){4-4} \cmidrule(r){5-5} \cmidrule(r){6-6}
Pythia-70m & 0.20 & 0.08 & 0.21 & 0.47 & $-$0.28 \\
Pythia-160m & $-$0.01 & 0.07 & 0.15 & 0.09 & $-$0.04 \\
Pythia-410m & 0.39 & 0.05 & 0.23 & 0.49 & $-$0.22 \\
Pythia-1b & $-$0.17 & 0.22 & 0.90 & 0.41 & $-$0.17 \\
Pythia-1.4b & 0.35 & 0.00 & 0.01 & 0.03 & $-$0.38 \\
Pythia-2.8b & 0.00 & $-$0.03 & 0.26 & 0.44 & $-$0.21 \\
Pythia-6.9b & 0.24 & $-$0.36 & $-$0.48 & $-$0.39 & $-$0.50 \\
Pythia-12b & 0.26 & $-$0.36 & $-$0.14 & $-$0.26 & $-$0.44 \\
\bottomrule
\end{tabular}
\caption{Estimated slowdown by JS-divergence}
\label{tab:effect_js}
\end{table}

\begin{table}
\centering
\footnotesize
\tabcolsep=0.01cm
\begin{tabular}{lllllllllll}
\toprule
 & \multicolumn{2}{c}{MVRR} & \multicolumn{2}{c}{NPS} & \multicolumn{2}{c}{NPZ} & \multicolumn{2}{c}{RC} & \multicolumn{2}{c}{Att.} \\
 \cmidrule(r){2-3} \cmidrule(r){4-5} \cmidrule(r){6-7} \cmidrule(r){8-9} \cmidrule(r){10-11}
Model  & \multicolumn{1}{c}{All} & \multicolumn{1}{c}{\textsc{RoI}} & \multicolumn{1}{c}{All} & \multicolumn{1}{c}{\textsc{RoI}} & \multicolumn{1}{c}{All} & \multicolumn{1}{c}{\textsc{RoI}} & \multicolumn{1}{c}{All} & \multicolumn{1}{c}{\textsc{RoI}} & \multicolumn{1}{c}{All} & \multicolumn{1}{c}{\textsc{RoI}} \\
\midrule
G2-sm & 31.9$^\dagger$ & 0.5 & 2.2 & 3.7 & 1.7 & 4.2$^\dagger$ & 24.7$^\dagger$ & 0.7 & 23.5$^\dagger$ & 3.6 \\
G2-md & 33.2$^\dagger$ & 2.1 & 4.4$^\dagger$ & 4.6$^\dagger$ & 6.3$^\dagger$ & 6.1$^\dagger$ & 30.5$^\dagger$ & 0.4 & 33.0$^\dagger$ & 5.8$^\dagger$ \\
G2-lg & 34.8$^\dagger$ & 3.9 & 3.3 & 2.2 & 4.5$^\dagger$ & 3.0 & 35.0$^\dagger$ & 0.6 & 33.9$^\dagger$ & 4.5$^\dagger$ \\
G2-xl & 31.4$^\dagger$ & 2.2 & 4.2$^\dagger$ & 1.0 & 5.5$^\dagger$ & 2.5 & 36.0$^\dagger$ & 0.4 & 33.7$^\dagger$ & 4.3$^\dagger$ \\
\cmidrule(r){1-1} \cmidrule(r){2-3} \cmidrule(r){4-5} \cmidrule(r){6-7} \cmidrule(r){8-9} \cmidrule(r){10-11}
OT-125m & 46.0$^\dagger$ & 0.2 & 3.2 & 1.4 & 1.9 & 5.4$^\dagger$ & 24.8$^\dagger$ & 0.2 & 28.2$^\dagger$ & 3.8 \\
OT-1.3b & 32.4$^\dagger$ & 2.7 & 5.8$^\dagger$ & 0.9 & 6.6$^\dagger$ & 2.3 & 29.2$^\dagger$ & 0.6 & 30.8$^\dagger$ & 3.9 \\
OT-2.7b & 33.0$^\dagger$ & 2.3 & 5.2$^\dagger$ & 0.9 & 7.8$^\dagger$ & 2.7 & 30.5$^\dagger$ & 0.9 & 29.6$^\dagger$ & 4.1$^\dagger$ \\
OT-6.7b & 28.6$^\dagger$ & 3.4 & 4.6$^\dagger$ & 2.8 & 7.0$^\dagger$ & 4.2$^\dagger$ & 29.9$^\dagger$ & 0.5 & 27.9$^\dagger$ & 3.1 \\
OT-13b & 31.4$^\dagger$ & 4.0$^\dagger$ & 5.0$^\dagger$ & 0.6 & 7.3$^\dagger$ & 3.2 & 29.1$^\dagger$ & 0.2 & 28.7$^\dagger$ & 3.3 \\
OT-30b & 31.0$^\dagger$ & 9.5$^\dagger$ & 3.8 & 3.3 & 6.0$^\dagger$ & 5.0$^\dagger$ & 25.4$^\dagger$ & 0.3 & 27.6$^\dagger$ & 3.3 \\
OT-66b & 28.5$^\dagger$ & 2.4 & 2.7 & 0.3 & 2.9 & 4.9$^\dagger$ & 21.5$^\dagger$ & 1.7 & 21.6$^\dagger$ & 4.1$^\dagger$ \\
\cmidrule(r){1-1} \cmidrule(r){2-3} \cmidrule(r){4-5} \cmidrule(r){6-7} \cmidrule(r){8-9} \cmidrule(r){10-11}
PT-70m & 17.5$^\dagger$ & 0.4 & 1.6 & 1.2 & 1.7 & 1.2 & 21.0$^\dagger$ & 1.6 & 24.8$^\dagger$ & 7.0$^\dagger$ \\
PT-160m & 15.4$^\dagger$ & 0.2 & 2.0 & 5.3$^\dagger$ & 3.7 & 2.2 & 23.6$^\dagger$ & 2.6 & 30.8$^\dagger$ & 6.1$^\dagger$ \\
PT-410m & 30.3$^\dagger$ & 0.3 & 3.3 & 0.7 & 1.5 & 2.9 & 27.2$^\dagger$ & 3.3 & 28.2$^\dagger$ & 4.8$^\dagger$ \\
PT-1b & 29.5$^\dagger$ & 1.1 & 3.6 & 2.1 & 3.3 & 5.9$^\dagger$ & 31.6$^\dagger$ & 2.9 & 31.6$^\dagger$ & 6.5$^\dagger$ \\
PT-1.4b & 38.1$^\dagger$ & 1.0 & 3.4 & 4.5$^\dagger$ & 1.9 & 2.9 & 25.6$^\dagger$ & 2.6 & 24.5$^\dagger$ & 5.0$^\dagger$ \\
PT-2.8b & 26.5$^\dagger$ & 0.1 & 3.3 & 2.2 & 3.9$^\dagger$ & 2.7 & 32.3$^\dagger$ & 2.7 & 29.5$^\dagger$ & 6.4$^\dagger$ \\
PT-6.9b & 29.8$^\dagger$ & 2.2 & 4.4$^\dagger$ & 1.3 & 2.9 & 1.2 & 30.0$^\dagger$ & 3.4 & 25.6$^\dagger$ & 6.7$^\dagger$ \\
PT-12b & 34.2$^\dagger$ & 0.9 & 3.2 & 1.4 & 2.9 & 2.9 & 28.9$^\dagger$ & 3.2 & 25.2$^\dagger$ & 5.1$^\dagger$ \\
\bottomrule
\end{tabular}
\caption{$\Delta$LL between Surprisal vs. Surprisal+JS settings. $^\dagger$ indicates statistical significance ($p < 0.05$) with a likelihood ratio test.}
\label{tab:surprisal_plus_js}
\end{table}

\end{document}